\documentclass[letterpaper,journal]{IEEEtran}

\usepackage{amsmath,amsfonts}
\usepackage{algorithmic}
\usepackage{algorithm}
\usepackage{array}
\usepackage{textcomp}
\usepackage{stfloats}
\usepackage{url}
\usepackage{verbatim}
\usepackage{graphicx}
\usepackage{cite}
\usepackage{xcolor}

\usepackage[hidelinks]{hyperref}

\DeclareFontFamily{T1}{formata}{}

\providecommand{\figcapfont}{\normalfont\footnotesize}
\definecolor{TUMBlue}{RGB}{0,101,189}%
\definecolor{TUMWhite}{RGB}{255,255,255}%
\definecolor{TUMBlack}{RGB}{0,0,0}%
\definecolor{TUMBlue1}{RGB}{0,51,89}%
\definecolor{TUMBlue2}{RGB}{0,82,147}%
\definecolor{TUMGray1}{RGB}{51,51,51}%
\definecolor{TUMGray2}{RGB}{127,127,127}%
\definecolor{TUMGray3}{RGB}{204,204,204}%
\definecolor{TUMBlue3}{RGB}{100,160,200}%
\definecolor{TUMBlue4}{RGB}{152,198,234}%
\definecolor{TUMIvory}{RGB}{218,215,203}%
\definecolor{TUMOrange}{RGB}{227,114,34}%
\definecolor{TUMGreen}{RGB}{162,173,0}%
\definecolor{TUMYellow}{RGB}{254,215,2}        %
\definecolor{TUMYellowDark}{RGB}{203,171,1}    %
\definecolor{TUMYellow1}{RGB}{254,222,52}      %
\definecolor{TUMYellow2}{RGB}{254,230,103}     %
\definecolor{TUMYellow3}{RGB}{254,238,154}     %
\definecolor{TUMYellow4}{RGB}{254,246,205}     %

\definecolor{TUMRed}{RGB}{234,114,55}        %
\definecolor{TUMRedDark}{RGB}{217,81,23}     %
\definecolor{TUMRed1}{RGB}{239,144,103}      %
\definecolor{TUMRed2}{RGB}{243,178,149}      %
\definecolor{TUMRed3}{RGB}{246,194,172}      %
\definecolor{TUMRed4}{RGB}{251,234,218}      %

\usepackage{tikz}
\usepackage{tikz-3dplot}
\usepackage{pgfplots}
\pgfplotsset{compat=1.18}
\usepackage[per-mode=fraction]{siunitx}
\usepackage[inkscapelatex=false, inkscapepath=./build/svg-inkscape]{svg}
\usepackage{placeins}
\usepackage{comment}
\usepackage{booktabs}
\usepackage{multirow}
\usepackage[capitalize, nameinlink]{cleveref}
\DeclareFontShape{T1}{formata}{m}{sl}{<->ssub*formata/m/it}{}
\DeclareFontShape{T1}{formata}{b}{sl}{<->ssub*formata/b/it}{}

\usepackage{mathtools}

\usetikzlibrary{external}
\ifdefined\OverleafCompilation
    \def\tikzexternalize{}

    \usepackage[placement=bottom,scale=3, contents=Overleaf Mode, opacity=0.2]{background}
\else
\fi

\usepgfplotslibrary{groupplots}
\usetikzlibrary{positioning}
\usetikzlibrary{shapes}
\usetikzlibrary{calc}
\usetikzlibrary{arrows.meta}
\usetikzlibrary{decorations.pathreplacing}
\usetikzlibrary{patterns}

\newcounter{subfigure}[figure]

\newcommand{\subfigcaptioncustom}[1]{
    {\figcapfont\par\scriptsize(\alph{subfigure})\;
            #1}
}
\newcommand{\startsubfigcustom}{\refstepcounter{subfigure}}
\newenvironment{subfigurecustom}
{%
    \refstepcounter{figure}%
    \setcounter{subfigure}{0}%
}
{%
    \addtocounter{figure}{-1}%
}

\newcommand{\githuburl}{\href{https://github.com/TUMFTM/3d-road-geometry-coupling}{github.com/TUMFTM/3d-road-geometry-coupling}}

\newcommand{\inputtikzfig}[1]{
    \includegraphics{./figures/tikz/#1.pdf}
}

\newcommand\trackflat{\ensuremath{\mathcal{T}_\mathrm{flat}}}
\newcommand\trackelevated{\ensuremath{\mathcal{T}_\mathrm{elevated}}}
\newcommand\trackbanked{\ensuremath{\mathcal{T}_\mathrm{banked}}}
\newcommand\trackverticalbank{\ensuremath{\mathcal{T}_\mathrm{vertical}}}

\tikzset{
node hyperlink/.style={
alias=sourcenode,
append after command={
let \n0={\pgfkeysvalueof{/pgf/outer xsep}},
\n1={\pgfkeysvalueof{/pgf/outer ysep}},
\p1=([shift={(\n0, -\n1)}] sourcenode.north west),
\p2=([shift={(-\n0, \n1)}] sourcenode.south east),
\n2={\x2-\x1},
\n3={\y1-\y2} in
node [inner sep=0pt, outer sep=0pt, anchor=north west, at=(\p1)] {\hyperref [#1]{\XeTeXLinkBox{\phantom{\rule{\n2}{\n3}}}}}
}
}
}

\pgfplotsset{
	compat=1.18,
	default-plot/.style={
			width=0.95\columnwidth,
			height=0.6\columnwidth,
			minor grid style={densely dotted, color=TUMGray3},
		},
	default-linestyle/.style={
			very thick,
			mark=none,
			mark size=1.0pt,
		},
	subplot_no_x_axis/.style={
			xticklabels=\empty, xlabel={}
		},
	every axis/.append style={
			label style={font=\footnotesize},      %
			tick label style={font=\footnotesize}, %
			legend style={font=\footnotesize},
			title style={font=\footnotesize}
		}
}

\pgfplotsset{
	validation_track_linestyles/.style={
			default-linestyle,
		},
	validaton_track_linestyles.banked/.style={
			validation_track_linestyles,
			color=TUMBlue,
			densely dashed
		},
	validaton_track_linestyles.elevated/.style={
			validation_track_linestyles,
			color=TUMOrange,
			densely dashdotted
		},
	validaton_track_linestyles.vertical/.style={
			validation_track_linestyles,
			color=black,
			densely dashdotted,
		},
	validaton_track_linestyles.flat/.style={
			validation_track_linestyles,
			color=TUMGreen,
			densely dashed
		},
}
\providecommand{\framevehicle}[1]{\prescript{}{\mathcal{V}}{#1}}
\providecommand{\frameinertial}[1]{\prescript{}{\mathcal{I}}{#1}}

\providecommand{\frameroad}[1]{\prescript{}{\mathcal{R}}{#1}}

\providecommand{\symbolroadplane}[1]{\mathord{\overline{#1}}}

\renewcommand{\cos}[1]{\mathrm{c}_{#1}}
\renewcommand{\sin}[1]{\mathrm{s}_{#1}}

\begin{document}

\title{Beyond the Plane: Coupling Planar Vehicle Dynamics with Three-Dimensional Road Geometry}

\author{Simon~Sagmeister,
    Phillip~Pitschi,
    Nico~Haja,
    and~Markus~Lienkamp%
    \thanks{S. Sagmeister and M. Lienkamp are with the Technical University of Munich, Germany; School of Engineering \& Design, Department of Mobility Systems Engineering, Institute of Automotive Technology.}%
    \thanks{P. Pitschi is with the Technical University of Munich, Germany; School of Engineering \& Design, Department of Engineering Physics \& Computation, Institute of Automatic Control.}%
    \thanks{Corresponding author: Simon Sagmeister (simon.sagmeister@tum.de).}%
    \thanks{This work was funded by the Deutsche Forschungsgemeinschaft (DFG, German Research Foundation) - 469341384}}

\markboth{Sagmeister \MakeLowercase{\textit{et al.}}: Beyond the Plane: Coupling Planar Vehicle Dynamics with Three-Dimensional Road Geometry}%
{Sagmeister \MakeLowercase{\textit{et al.}}: Beyond the Plane: Coupling Planar Vehicle Dynamics with Three-Dimensional Road Geometry}

\maketitle

\begin{abstract}
    Simulation is crucial for developing and testing autonomous driving systems.
    In particular, the development of localization and control algorithms relies on an accurate vehicle dynamics simulation.
    However, most vehicle dynamics models are two-dimensional while real-world roads are three-dimensional.
    For example, effects from the three-dimensional road geometry on the Las Vegas Motor Speedway can increase the normal forces on the tires by more than \SI{66}{\percent} compared to the nominal load at standstill.
    As a result, even highly detailed planar vehicle dynamics models struggle to accurately reproduce the real vehicle's behavior.
    While solutions for three-dimensional vehicle dynamics exist, they are rarely adopted, computationally expensive, and complex.
    To address this issue, we present a novel method to couple planar vehicle dynamics models with real-world three-dimensional road geometry.
    We transform the planar vehicle state from the vehicle model's two-dimensional plane to its corresponding representation in three-dimensional space.
    Additionally, we calculate road-geometry-induced forces and moments and apply them to the planar vehicle model.
    We validate our approach using high-speed data recorded with a full-scale race car on the banked Las Vegas Motor Speedway.
    Furthermore, on synthetic tracks, we show that our method yields accurate results even in edge cases.
    Together, our results demonstrate that the gap between planar simulation and real-world three-dimensional roads can be closed without abandoning simpler planar models.
    To simplify adoption of our method, we provide the implementation as open-source software on \mbox{\githuburl}.
\end{abstract} 

\section{Introduction}
\label{sec:introduction}

\IEEEPARstart{S}{imulation} is a key element in the development of autonomous driving functions and algorithms. Especially for state estimation and vehicle dynamics control, simulation is indispensable, since real-world testing is expensive and time-consuming. This is further exacerbated in use cases where no safety driver can be present, such as autonomous racing. In such settings, deploying new untested algorithms can lead to crashes and severe damage to the test vehicle.

\begin{figure}[!tb]
    \centering
    \includegraphics[width=\columnwidth]{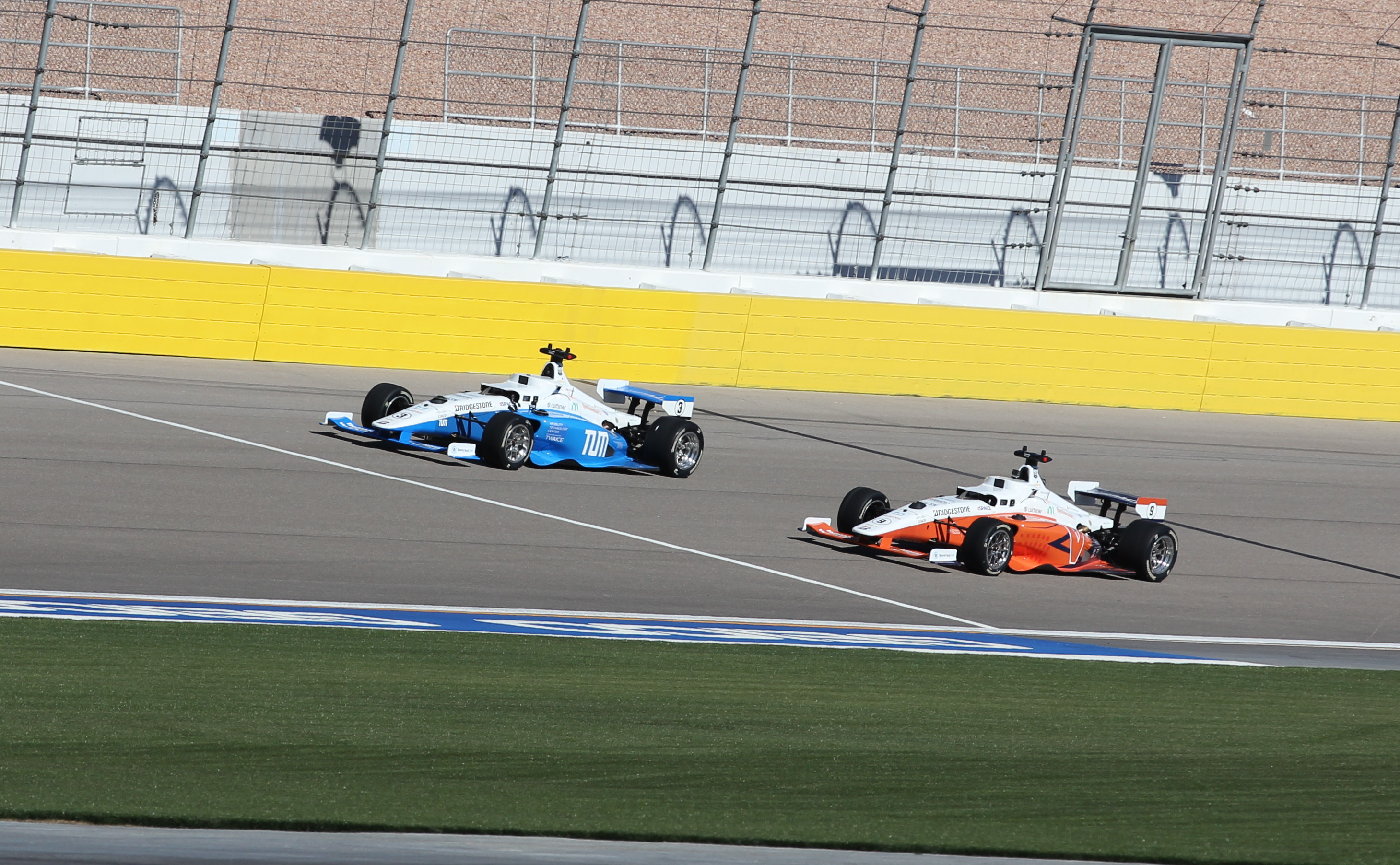}
    \caption{Our AV21 race car used for data collection during an overtaking maneuver on the banked Las Vegas Motor Speedway.}
    \label{fig:vehicle_on_banked_road}
\end{figure}

Virtual testing and development require a simulation that accurately represents the real-world behavior of the test vehicle. However, while real-world dynamics are three-dimensional, vehicle dynamics simulations often rely on simplified two-dimensional models~\cite{zeng2025comparativestudytrajectory, calzolari2017Comparisontrajectorytracking, wischnewski2022Tubemodelpredictive,sagmeister2024AnalyzingImpactSimulation, liniger2015OptimizationBasedAutonomousRacing}.
Even on seemingly flat surfaces, this can lead to discrepancies between simulation and reality.
For example, on the Formula~1 race track in Monza, the Intertial Measurement Unit (IMU) of our autonomous race car (Fig.~\ref{fig:vehicle_on_banked_road}) measured a normal vertical acceleration of up to \SI{11.5}{\meter\per\second\squared} when driving at an apex velocity of \SI{29.5}{\meter\per\second} through the ``Lesmo~2'' turn. This increases the normal load on the tires by more than \SI{17}{\percent} over the nominal load at standstill, significantly influencing the vehicle's dynamic response. In a more extreme example, when driving at a constant velocity of \SI{69.5}{\meter\per\second} on the banked Las Vegas Motor Speedway, the IMU measurement reveals a vertical acceleration of up to \SI{16.3}{\meter\per\second\squared}, corresponding to an increase in normal load of more than \SI{66}{\percent} compared to standstill.

While solutions for correctly modeling three-dimensional vehicle dynamics exist, they are often complex, computationally expensive, and not openly available (Section~\ref{sec:related_work}).
Moreover, the required increase in complexity for building three-dimensional models heavily obscures the underlying vehicle dynamics modeled.
This compromises the understandability and interpretability of simulation results, which is particularly problematic for the development of motion estimation and control algorithms.
In contrast, most publications still rely on planar vehicle dynamics models, which are simpler to interpret, but fail to capture the effects from the three-dimensional road geometry~\cite{wischnewski2022Tubemodelpredictive,liniger2015OptimizationBasedAutonomousRacing,thrun2006StanleyRobotthat, calzolari2017Comparisontrajectorytracking}.

To address these issues, we propose a method that utilizes existing planar vehicle dynamics models for simulations on three-dimensional roads.
Our method does not modify the underlying planar vehicle dynamics model; instead, it couples the model with the real-world road geometry. This allows us to retain the simplicity and interpretability of planar vehicle dynamics models while still simulating on three-dimensional roads. To validate our approach, we use both synthetic reference tracks and real-world data. Fig.~\ref{fig:vehicle_on_banked_road} shows our AV21 race car used for data collection during an Indy Autonomous Challenge \cite{mitchell2024IndyAutonomousChallenge} event on the banked Las Vegas Motor Speedway.
In summary, this work comprises the following main contributions:
\begin{itemize}
    \item A novel method for creating vehicle dynamics simulations on three-dimensional roads by coupling existing black-box planar vehicle dynamics models with the real-world road geometry.
    \item Validation of our method using synthetically generated reference tracks and real-world data recorded on the banked Las Vegas Motor Speedway.
    \item An open-source implementation of our method that can be easily integrated with existing vehicle models and simulation frameworks. It is available under \mbox{\githuburl}.
\end{itemize}

\section{Related Work}
\label{sec:related_work}

Previous work has incorporated effects from three-dimensional road geometry into vehicle dynamics models. We cluster these works into three categories.
First, we present approaches that adapt autonomous driving algorithms to three-dimensional roads.
Second, we discuss commercial high-fidelity simulators before discussing open-source simulators and vehicle models.
Finally, we identify a research gap in the existing literature that motivates our work.

\subsection{Autonomous Driving Algorithms}

Autonomous driving algorithms were previously adapted to operate on three-dimensional road geometries.
This mainly includes raceline optimization \cite{rowold2023OnlineTimeOptimalTrajectory,limebeer2015OptimalControlFormula,lovato2021Curvedribbonbasedtrackmodelling}, trajectory planning \cite{ogretmen2024SamplingBasedMotionPlanning}, trajectory tracking control \cite{ruifeng2026Modellingcontrolhighsideslip,fork2024Noneuclideanvehiclemotion,fork2021ModelsPredictiveControl}, and state estimation algorithms \cite{goblirsch2024ThreeDimensionalVehicleDynamics, dahmani2013Vehicledynamicestimation}.
The level of fidelity when incorporating three-dimensional road geometry varies from simpler approaches considering the change in orientation of the gravitational force~\cite{zerbato2024BifurcationAnalysisNonlinear,raji2022Motionplanningcontrol,wischnewski2022TubeMPCApproachAutonomous} to full three-dimensional formulations of the underlying dynamics model~\cite{bongard2026RobustNonlinearTrajectory, fork2021ModelsPredictiveControl, fork2024Noneuclideanvehiclemotion,limebeer2015OptimalControlFormula, baxter2026HighSpeedAllTerrainAutonomy}.
These models are often formulated in curvilinear coordinates along a reference line~\cite{bongard2026RobustNonlinearTrajectory,raji2022Motionplanningcontrol}, though some works also use a global coordinate system~\cite{fork2021ModelsPredictiveControl, fork2024Noneuclideanvehiclemotion,goblirsch2024ThreeDimensionalVehicleDynamics}.

Further, several publications adapt the planar vehicle equations commonly found in textbooks such as~\cite{milliken1995Racecarvehicle,allen1992Vehicledynamicstability,gillespie1992FundamentalsVehicleDynamics} to three-dimensional road surfaces.
These often propose models used in an optimal control context, which they then also use for simulation~\cite{fork2021ModelsPredictiveControl,rowold2023OnlineTimeOptimalTrajectory,limebeer2015OptimalControlFormula}.
Fork et al.~\cite{fork2024Noneuclideanvehiclemotion,fork2021ModelsPredictiveControl,fork2024Modelsgroundvehicle,fork2024General3DRoad} present methods for creating three-dimensional vehicle models on non-planar roads by combining different road surface representations with vehicle models of varying detail. The resulting equations of motion are derived using a symbolic math toolbox~\cite{fork2024Noneuclideanvehiclemotion}. However, especially when coupled with more complex vehicle dynamics models, this can result in lengthy, hard-to-interpret equations.
Limebeer et al.~\cite{limebeer2015OptimalControlFormula,perantoni2015OptimalControlFormula} present a method for modeling three-dimensional road surfaces using a ribbon approach. This enables the derivation of equations of motion for a vehicle model on three-dimensional roads in curvilinear coordinates along a reference line.
Rowold et al.~\cite{rowold2023OnlineTimeOptimalTrajectory} simplify this model for computational efficiency, enabling its use in an online trajectory optimization framework. However, this simplified model lacks the detail required for vehicle dynamics simulation.

The existence of these approaches indicates the need for incorporating three-dimensional road geometry into the development of autonomous driving algorithms. However, while these approaches are mathematically rigorous, they often require a complete reformulation of the underlying vehicle dynamics model.

\subsection{Commercial High-Fidelity Simulators}

In academia and industry, commercial multi-body vehicle dynamics simulators are widely used to develop physically accurate, complex models of a test vehicle~\cite{silva2024Realistic3DSimulators, kaur2021SurveySimulatorsTestinga}. Commercial software packages such as CarSim~\cite{carsimTerrainRoadSurfaces, sayers1996GenericMultibodyVehicle, sayers2011RoadCharacterizationSimulation} offer high-fidelity three-dimensional modeling, allowing the road surface to be explicitly defined along a reference path. Similarly, there are other closed-source tools such as Modelica's Vehicle Dynamics Library~\cite{andreasson2011VehicleDynamicsLibrary} and the dSPACE ASM~\cite{patil2012HybridVehicleModel}, which provide comprehensive simulation frameworks.

In parallel, racing games and their underlying physics engines have gained traction for autonomous driving research~\cite{li2024ChooseYourSimulator}. High-fidelity titles such as Assetto Corsa~\cite{bertogna2024SimulationBenchmarkAutonomous} and Gran Turismo~\cite{wurman2022OutracingchampionGran} feature sophisticated three-dimensional simulation engines that natively handle complex multi-body phenomena on non-planar tracks.
While these specialized commercial simulators and racing engines capture complex 3D geometry and uneven road interactions~\cite{Varunjikar2011MultibodyVD, tengler2021EffectiveAlgorithmUneven}, their internal physics engines are proprietary. Furthermore, their computational overhead and opaque black-box formulations make them ill-suited for rapid algorithm design, mathematical interpretability, and parameter tuning required in scientific algorithm development.

\subsection{Open-Source Vehicle Simulation}

While open-source simulators such as CARLA~\cite{dosovitskiy2017CARLAopenurban} and R-CARLA~\cite{brunner2025RCARLAHighFidelitySensor}, or the Chrono multibody physics engine~\cite{tasora2016ChronoOpenSource,serban2019ChronoVehicletemplatebased}, have similar drawbacks in terms of computational efficiency, they are more transparent because their source code is publicly available. However, they still rely on complex multi-body physics, which complicates the interpretability of the underlying vehicle dynamics, and struggle with determinism~\cite{chance2022DeterminismGameEngines}.

\subsection{Research Gap}

As the existing literature demonstrates, there currently is a fundamental tradeoff in vehicle dynamics simulation for autonomous driving.
At one end of the spectrum, high-fidelity multi-body simulators natively handle spatial three-dimensional tracks but suffer from significant computational overhead, limited determinism, and opaque, proprietary models.
At the other end, analytical planar models offer high computational efficiency and crucial mathematical transparency for algorithm development, but they inherently fail to capture the complex spatial dynamics induced by banked curves and varying elevations.
While custom non-planar mathematical models exist, they demand tedious, use-case-specific reformulations of the fundamental differential equations, often resulting in overwhelmingly complex expressions. %
\section{Methodology}
\label{sec:methodology}

This section presents our approach for adapting a planar vehicle model to three-dimensional road geometries. After introducing the necessary preliminaries and notation, we describe the transformation of geometric quantities and measurements from the two-dimensional road plane into three-dimensional space. Finally, we present the calculation of the road forces and moments acting on the vehicle due to the road geometry.

\subsection{Preliminaries}

We employ right-handed Cartesian coordinate frames, using a leading subscript to identify the specific reference frame of a given quantity. The vehicle frame is denoted by $\framevehicle{\square}$ and the inertial frame by $\frameinertial{}{\square}$. Quantities in the velocity frame are expressed without a leading subscript. Within the inertial frame, the $\frameinertial{x}$-, $\frameinertial{y}$-, and $\frameinertial{z}$-axes are defined to point in the North, East, and Up directions, respectively.

Measurements obtained from the planar vehicle model are denoted as $\symbolroadplane{\square}$, whereas values without such marking refer to the three-dimensional representation. Derivatives with respect to time are denoted as $\frac{\mathrm{d}\square}{\mathrm{d}t} = \dot{\square}$ and with respect to the arc length $s$ as $\frac{\mathrm{d}\square}{\mathrm{d}s} = \square^\prime$. To shorten notation, cosine and sine are abbreviated by $\mathrm{cos}(\square) = \cos{\square}$ and $\mathrm{sin}(\square) = \sin{\square}$, respectively.

The transformation of quantities from the velocity frame to the vehicle frame is achieved using the following rotation matrix:
\begin{equation}
    \boldsymbol{R}_z(\beta) =
    \begin{bmatrix}
        \cos{\beta} & -\sin{\beta} & 0 \\
        \sin{\beta} & \cos{\beta}  & 0 \\
        0           & 0            & 1
    \end{bmatrix}
\end{equation}
where $\beta = \arctan\left(\frac{\framevehicle{v}_y}{\framevehicle{v}_x}\right)$ is the sideslip angle.

For the planar vehicle model, we assume the angular velocities with respect to the arc length $\symbolroadplane{\Omega}_x$ and $\symbolroadplane{\Omega}_y$, the angular velocities $\symbolroadplane{\omega}_x$ and $\symbolroadplane{\omega}_y$, the angular accelerations with respect to the arc length $\symbolroadplane{\Omega}'_x$ and $\symbolroadplane{\Omega}'_y$, the angular accelerations $\dot{\symbolroadplane{\omega}}_x$ and $\dot{\symbolroadplane{\omega}}_y$, and the banking and slope angles $\symbolroadplane{\varphi}$ and $\symbolroadplane{\mu}$ to be zero:
\begin{equation}
    \begin{aligned}
        \symbolroadplane{\Omega}_x       & \stackrel{!}{=} 0
                                         & \qquad
        \symbolroadplane{\Omega}_y       & \stackrel{!}{=} 0
        \\
        \symbolroadplane{\omega}_x       & \stackrel{!}{=} 0
                                         & \qquad
        \symbolroadplane{\omega}_y       & \stackrel{!}{=} 0
        \\
        \symbolroadplane{\Omega}'_x      & \stackrel{!}{=} 0
                                         & \qquad
        \symbolroadplane{\Omega}'_y      & \stackrel{!}{=} 0
        \\
        \dot{\symbolroadplane{\omega}}_x & \stackrel{!}{=} 0
                                         & \qquad
        \dot{\symbolroadplane{\omega}}_y & \stackrel{!}{=} 0
        \\
        \symbolroadplane{\varphi}        & \stackrel{!}{=} 0
                                         & \qquad
        \symbolroadplane{\mu}            & \stackrel{!}{=} 0
    \end{aligned}
    \label{eq:assumptions}
\end{equation}

\subsection{Track Representation}

\begin{figure}[!tb]
    \centering
    \inputtikzfig{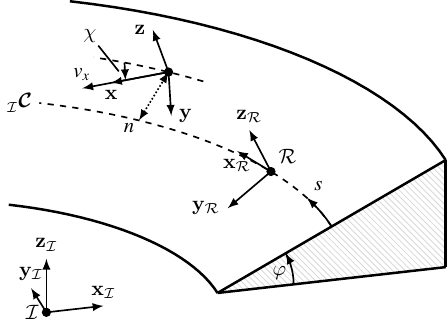}
    \caption{Relevant coordinate frames on the three-dimensional road surface. Modified from \cite{rowold2023OnlineTimeOptimalTrajectory}. The velocity frame is offset by $n$ and rotated by $\chi$ from the reference line $\frameinertial{\boldsymbol{\mathcal{C}}}$. The vehicle frame $\mathcal{V}$ is obtained by rotating the velocity frame by the sideslip angle $\beta$ around its $z$-axis.}
    \label{fig:track-representation}
\end{figure}

This work uses the ribbon-based track representation proposed by Perantoni et al. \cite{perantoni2015OptimalControlFormula}, which characterizes the track via a spine and a lateral offset. The spine is defined as a three-dimensional curve $\frameinertial{\mathcal{C}}=\{\frameinertial{p}(s)= \begin{bmatrix} p_x(s) & p_y(s) & p_z(s) \end{bmatrix}^T  \mid s \in \left[0, s_\mathrm{f}\right] \}$, parametrized by the arc length $s$ up to the total track length $s_\mathrm{f}$. As depicted for a banked left turn in Fig.~\ref{fig:track-representation}, the kinematic description relies on three principal reference frames. The road frame $\mathcal{R}$ originates on the spine. Its $\frameroad{x}$-axis is tangent to the path, the $\frameroad{y}$-axis lies in the road plane perpendicular to the tangent, and the $\frameroad{z}$-axis is the upward-pointing surface normal. The velocity frame is obtained by translating $\mathcal{R}$ by a lateral offset $n$ along the $\frameroad{y}$-axis and applying a rotation $\chi$ about the local $\frameroad{z}$-axis, with its origin lying on the road plane below the vehicle's center of gravity. Lastly, the vehicle frame $\mathcal{V}$ is derived from the velocity frame through a rotation by the sideslip angle $\beta$ about the $z$-axis, such that its $\framevehicle{x}$-axis aligns with the longitudinal axis of the vehicle.

\subsection{General Strategy}

\begin{figure}[!tb]
    \centering
    \inputtikzfig{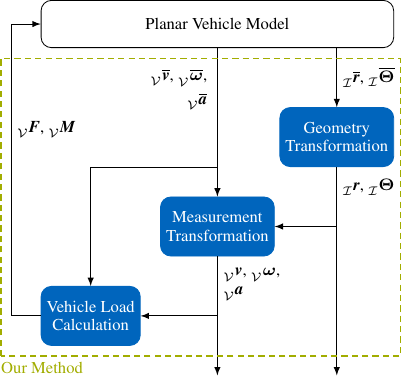}
    \caption{Outline of our approach. The planar vehicle model is treated as a black box. The modules drawn in blue comprise our method and are presented in the remainder of this section.}
    \label{fig:method-overview}
\end{figure}

After introducing the necessary preliminaries and the track representation, we now present the core concept of our approach. We present the high-level approach in this section, followed by a detailed description of the individual components in the remainder of the methodology section. The main idea behind our method is that we do not modify the planar vehicle model; instead, we create a separate module that adapts any black-box planar vehicle model to the three-dimensional road surface. This methodology is threefold and visualized in Fig.~\ref{fig:method-overview}.

Instead of modifying the model's internals, we interact solely with the vehicle model through its input and output interfaces. We use the planar outputs from the vehicle model and transform them according to the car's current state on the track to obtain the corresponding three-dimensional quantities. This is done in two distinct steps: First, we perform a geometry transformation to convert the vehicle's planar pose into a three-dimensional pose on the track. Second, we transform the remaining measurements from the planar vehicle model into their three-dimensional equivalents. Both of these steps are described in detail in Sections~\ref{sec:geo-trafo} and \ref{sec:meas-trafo}, respectively.
However, since the three-dimensional nature of the track also influences the vehicle model, we apply external forces and moments acting on the vehicle's center of gravity. The derivation of these forces is described in Section~\ref{sec:load-calc}.
This way, our method is compatible with any planar vehicle as long as it accepts external forces and moments as input.

Applying the road-induced forces and moments at the vehicle's center of gravity is an approximation, since in a full multi-body formulation, these loads would act on the individual bodies of the vehicle. By collapsing the effects onto the center of gravity, we omit secondary effects such as gyroscopic moments on the individual wheels. We justify this simplification since the sprung mass is much larger than the mass of the unsprung components~\cite{sagmeister2024OpenCarDynamics,milliken1995Racecarvehicle}. Our real-world results in Section~\ref{sec:results} confirm that this approximation is sufficient to reproduce the measured accelerations and rates on the banked Las Vegas Motor Speedway.

\subsection{Geometry Transformation}
\label{sec:geo-trafo}

A physical vehicle driving on a three-dimensional track always drives along its actual road surface. Since the road surface is not a flat plane, the pose of a planar vehicle model must be transformed to reflect the vehicle's actual three-dimensional pose on the track. This becomes apparent when considering a thought experiment of a \SI{90}{\degree} banked oval race track. A real-world vehicle would only need to follow a straight line to complete a lap around this race track. However, a planar vehicle that neglects all three-dimensional effects would instead need to follow the track's oval reference line. For the planar model to behave similarly, a geometry transformation is required.

\begin{figure}[!tb]
    \centering
    \vspace*{0.2cm}
    \centering
    \begin{subfigurecustom}
        \begin{minipage}[t]{0.49\columnwidth}
            \centering
            \startsubfigcustom
            \includegraphics[width=0.95\columnwidth]{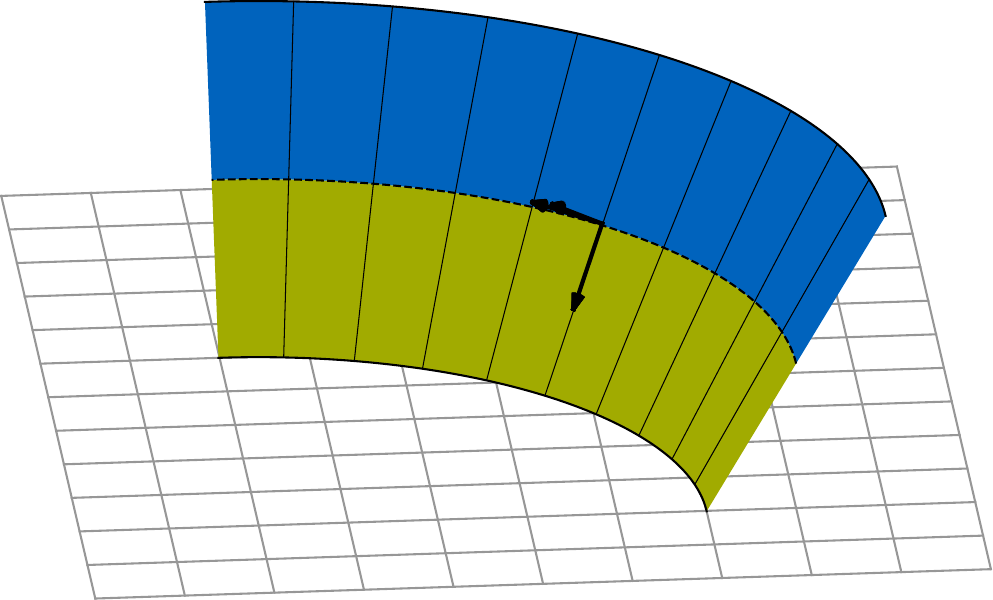}
            \subfigcaptioncustom{3D Road Geometry}
            \label{fig:winded-track}
        \end{minipage}
        \begin{minipage}[t]{0.49\columnwidth}
            \centering
            \startsubfigcustom
            \vspace{6.2mm}
            \includegraphics[width=0.95\columnwidth]{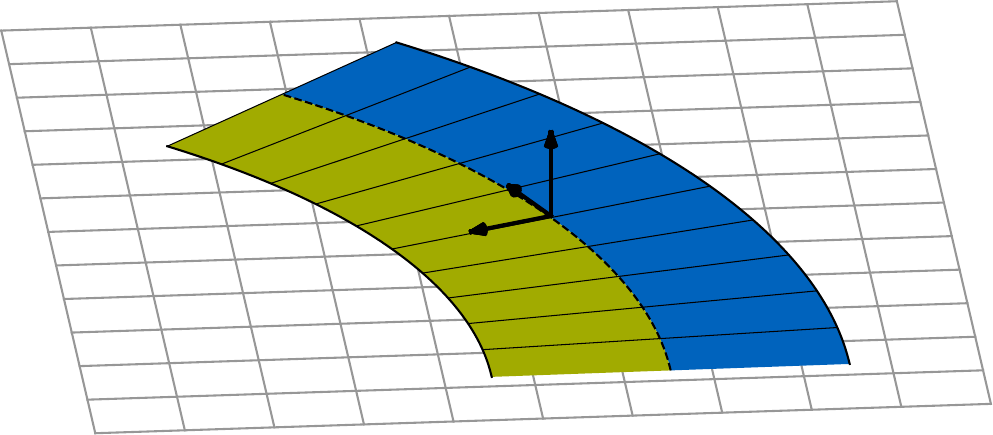}
            \subfigcaptioncustom{Flat Pattern of the Road Geometry}
            \label{fig:unwound-track}
        \end{minipage}
    \end{subfigurecustom}
    \par\vspace*{0.2cm}
    \caption{Conceptual illustration of the geometry transformation. The road surface is cut into small slices and laid flat on the ground.}
    \label{fig:geometry-transformation-idea}
\end{figure}

The core concept of this transformation is to project the three-dimensional track surface onto the two-dimensional plane on which the planar vehicle model is driving. Fig.~\ref{fig:geometry-transformation-idea} visualizes this concept. The three-dimensional road surface in Fig.~\ref{fig:winded-track} is projected onto a plane in Fig.~\ref{fig:unwound-track}.
This projection can be imagined as cutting the road surface perpendicular to a reference line into infinitesimally small pieces and reassembling them onto a plane.
For the remainder of this work, we refer to this two-dimensional embedding of the three-dimensional road surface as the road plane.

We design the transformation to preserve the local curvature along the road surface $\Omega_z$ and the length $s$ along the reference line. Therefore, we can express the reference line's curvature $\symbolroadplane{\Omega}_z$ and the arc length $\symbolroadplane{s}$ within the road plane as follows:
\begin{align}
    \symbolroadplane{\Omega}_z & = \Omega_z \\
    \symbolroadplane{s}        & = s
\end{align}

Since these quantities are identical across all representations in this work, we drop the road plane symbol and simply write $\Omega_z$ and $s$ in the following.
Since our track representation is given as arrays of numerical values rather than a continuous function, we construct the reference line within the road plane by numerically integrating the curvature ${\Omega}_z$ along the arc length $s$, given a starting pose.

\begin{figure}[!tb]
    \centering
    \inputtikzfig{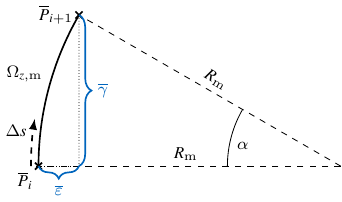}
    \caption{Circular integration scheme for improving the accuracy of numerical integration during construction of the planar reference line.}
    \label{fig:circular-integration}
\end{figure}

However, early experiments revealed that this does not reproduce the reference line of a flat track with sufficient accuracy.
The numerical integration error accumulates over the course of the track and primarily results from the Euler forward integration scheme used to construct the reference line.
For the flat reference track described in Section~\ref{sec:results}, this resulted in a final displacement of several meters compared to the reference line.
While a higher-order integrator could mitigate this issue, it is more efficient to leverage the fact that we have the curvature $\Omega_z$ available along the whole track.
Since Euler forward integration assumes a constant heading during the integration step, it effectively connects points with straight segments.
However, since we know the curvature $\Omega_z$, we instead use a circular arc to calculate the position of the next point $\symbolroadplane{P}_{i+1}$ given the current point $\symbolroadplane{P}_i$. Fig.~\ref{fig:circular-integration} visualizes this idea.

Since the curvature changes from point $\symbolroadplane{P}_i$ to $\symbolroadplane{P}_{i+1}$, we use the mean curvature $\Omega_{z,\mathrm{m}}$ during integration:
\begin{equation}
    \Omega_{z,\mathrm{m}} = \frac{\Omega_{z,i} + \Omega_{z,i+1}}{2}
\end{equation}

We can calculate the arc angle $\alpha$ between the two points from the mean radius $R_\mathrm{m}$ and the distance $\Delta{}s$ between them as follows:
\begin{equation}
    \alpha = \frac{\Delta{}s}{R_\mathrm{m}} = \Delta{}s\Omega_{z,\mathrm{m}}
\end{equation}

As depicted in Fig.~\ref{fig:circular-integration}, the lateral distance $\symbolroadplane{\varepsilon}$ and the longitudinal distance $\symbolroadplane{\gamma}$ between the two points can then be calculated using basic trigonometry:
\begin{align}
    \symbolroadplane{\varepsilon} & = R_\mathrm{m}(1 - \cos{\alpha}) =  \frac{(1 - \cos{\alpha})}{\Omega_{z,\mathrm{m}}}
    \label{eq:epsilon}                                                                                                   \\
    \symbolroadplane{\gamma}      & = R_\mathrm{m}\sin{\alpha}       = \frac{\sin{\alpha}}{\Omega_{z,\mathrm{m}}}
    \label{eq:gamma}
\end{align}

These equations become singular as $\Omega_z$ approaches zero. To address this, we apply a small-angle approximation for small curvatures to avoid numerical issues. This reveals that for small curvatures, the lateral distance $\symbolroadplane{\varepsilon}$ approaches zero, while the longitudinal distance $\symbolroadplane{\gamma}$ approaches the distance $\Delta{}s$ between the two points:

\begin{equation}
    \symbolroadplane{\varepsilon} \approx \frac{(1 - 1)}{\Omega_{z,\mathrm{m}}} = 0
\end{equation}
\begin{equation}
    \symbolroadplane{\gamma} \approx \frac{\alpha}{\Omega_{z,\mathrm{m}}}  = \frac{\Delta{}s\Omega_{z,\mathrm{m}}}{\Omega_{z,\mathrm{m}}} = \Delta{}s
\end{equation}

This means that the circular arc integration reduces to a conventional Euler forward integration scheme for small curvatures. This does not compromise numerical accuracy, since the Euler integration scheme is well-suited for small curvatures, as it connects points with straight segments. Using the heading $\symbolroadplane{\theta}_i$ at point $\symbolroadplane{P}_i$ together with the distances $\symbolroadplane{\varepsilon}$ and $\symbolroadplane{\gamma}$, we can calculate the position of the next point $\symbolroadplane{P}_{i+1}$:
\begin{equation}
    \begin{bmatrix}
        \symbolroadplane{x}_{i+1} \\
        \symbolroadplane{y}_{i+1}
    \end{bmatrix} = \begin{bmatrix}
        \symbolroadplane{x}_{i} \\
        \symbolroadplane{y}_{i}
    \end{bmatrix} +
    \begin{bmatrix}
        -\sin{{\symbolroadplane{\theta}_i}} & \cos{\symbolroadplane{\theta}_i} \\
        \cos{\symbolroadplane{\theta}_i}    & \sin{\symbolroadplane{\theta}_i}
    \end{bmatrix}
    \begin{bmatrix}
        \symbolroadplane{\varepsilon} \\
        \symbolroadplane{\gamma}
    \end{bmatrix}
\end{equation}

This circular integration scheme allows us to accurately reconstruct the reference line of a flat reference track.

\begin{figure}[!tb]
    \centering
    \inputtikzfig{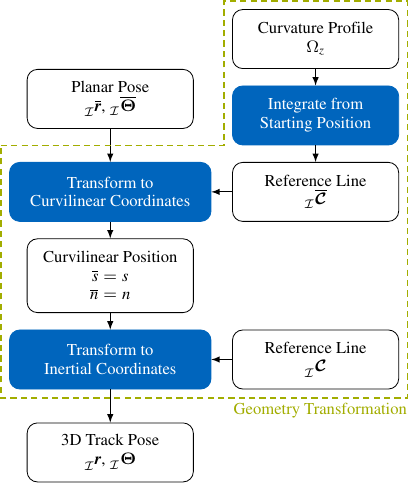}
    \caption{Overview of the geometry transformation for constructing a three-dimensional pose on the race track from the planar pose of the vehicle dynamics model.}
    \label{fig:flowchart-transformation-routine}
\end{figure}

Now that we have derived the scheme for calculating the reference line within the virtually constructed road plane, we can use this reference line to project the pose of the planar vehicle model into three-dimensional space. Fig.~\ref{fig:flowchart-transformation-routine} shows an overview of the transformation procedure.

We first transform the planar pose into curvilinear coordinates relative to the road plane reference line $\frameinertial{\boldsymbol{\symbolroadplane{\mathcal{C}}}}$ using an algorithm adapted from \cite{wursching2024RobustEfficientCurvilinear, bender2014LaneletsEfficientmap, pek2020CommonRoadDrivabilityChecker}.
These curvilinear coordinates are defined to be identical in both the road plane and the three-dimensional track representation, since the geometry transformation preserves the curvature and arc length along the reference line. By converting the curvilinear coordinates back into inertial coordinates using the three-dimensional reference line $\frameinertial{\boldsymbol{\mathcal{C}}}$, we obtain the three-dimensional position of the vehicle on the track. The following section provides a detailed description of how to obtain the correct three-dimensional orientation from this position and the relative orientation $\chi$.

One caveat of this method is that it significantly changes the geometry of the reference line. While race tracks are usually closed circuits, meaning that their start and end points coincide, this property may no longer hold when projecting the reference line onto the road plane. Furthermore, the resulting change in geometry could cause the reference line $\frameinertial{\symbolroadplane{\boldsymbol{\mathcal{C}}}}$ to self-intersect if it becomes sufficiently long. Such self-intersection must be avoided, since it would render the transformation of the planar pose into curvilinear coordinates ambiguous.

To mitigate this, we only calculate short segments (100 points, \SI{1}{\meter} spacing) of the road-plane reference line at a time, starting from the vehicle's current position. We iteratively update this line as the vehicle moves along the track, ensuring a seamless transition at the joints between the segments. By creating this short-term reference line, we can avoid self-intersection and ensure that the transformation from the planar pose to curvilinear coordinates remains well-defined.

\subsubsection{Orientation Transformation to Inertial Coordinates}
\label{sec:orientation-trafo}

To obtain the vehicle orientation in the inertial frame, a transformation from the curvilinear coordinates associated with the reference line $\frameinertial{\mathcal{C}}$ is required. The absolute orientation combines the orientation of the reference line at the current arc length $s$, characterized by heading $\theta$, banking $\varphi$, and slope $\mu$, with the relative vehicle orientation in the road plane given by $\chi - \beta$. The combined rotation is expressed as a transformation sequence from the inertial frame to the vehicle frame. We decompose this rotation into a sequence from the inertial to the road frame using the banking, slope, and heading angles, followed by an in-plane rotation by the angle $\chi - \beta$ that resolves the road frame vector into the vehicle frame:

\begin{equation}
    \boldsymbol{R}_{\mathcal{I}\mathcal{V}} = \boldsymbol{R}_{z}(\theta)  \boldsymbol{R}_{y}(\mu)  \boldsymbol{R}_{x}(\varphi)  \boldsymbol{R}_{z}(\chi-\beta)
\end{equation}

\pagebreak
We then extract the Euler angles from the entries $r_{ij}$ of the resulting rotation matrix $\boldsymbol{R}_{\mathcal{I}\mathcal{V}}$:
\begin{align}
                                                  & \mu = \arctan2\left(-r_{31},\sqrt{r_{11}^2 + r_{21}^2}\right) \\
    \mathrm{if}~\mu \neq \pm \frac{\pi}{2}: \quad & \theta = \arctan2\left(r_{21},r_{11}\right),                  \\
                                                  & \varphi = \arctan2\left(r_{32},r_{33}\right)                  \\
    \mathrm{else}: \quad                          & \theta = \arctan2\left(-r_{12},r_{22}\right),                 \\
                                                  & \varphi = 0
\end{align}

\subsection{Measurement Transformation}
\label{sec:meas-trafo}

The transformed measurements must accurately reflect the quantities that the vehicle's onboard sensors would register on a three-dimensional track. The outputs of the planar vehicle model, however, inherently lack the effects induced by three-dimensional road geometry and gravitational forces. These measurements must therefore be transformed to represent their true three-dimensional equivalents. For each sensor measurement, we derive expressions for the spatial quantities in the velocity frame, identify the corresponding subset modeled by the planar vehicle model, and apply a transformation to yield measurements resolved in the vehicle frame.

\subsubsection{Linear Velocity Transformation}

We derive the three-dimensional representation of the velocity by expanding the expression used by Rowold et al. \cite{rowold2023OnlineTimeOptimalTrajectory} with an additional heaving motion term $\dot{h}$:
\begin{equation}
    \boldsymbol{v} = \begin{bmatrix}
        v_x \\ 0 \\ w + \dot{h}
    \end{bmatrix} + \boldsymbol{\omega} \times \begin{bmatrix}
        0 \\ 0 \\ h
    \end{bmatrix} = \begin{bmatrix}
        v_x + \omega_yh \\
        -\omega_xh      \\
        w + \dot{h}
    \end{bmatrix}
    \label{eq:vel-general}
\end{equation}

The total velocity in the road plane is denoted by $v_x$, the vertical velocity perpendicular to the road plane by $w$, and the height of the vehicle's center of gravity with respect to the road plane by $h$.

Evaluating this equation under the assumptions for the planar vehicle model in \eqref{eq:assumptions} and using the relationship $\symbolroadplane{w} = n \symbolroadplane{\omega}_x$ isolates the contribution modeled by the planar simulation:
\begin{equation}
    \symbolroadplane{\boldsymbol{v}} =
    \begin{bmatrix}
        \symbolroadplane{v}_x \\
        \symbolroadplane{v}_y \\
        \symbolroadplane{v}_z
    \end{bmatrix}
    \stackrel{!}{=}
    \begin{bmatrix}
        v_x \\
        0   \\
        \dot{h}
    \end{bmatrix}
    \label{eq:vel-equality}
\end{equation}

Substituting \eqref{eq:vel-equality} into \eqref{eq:vel-general} yields the velocity transformation formula:

\begin{equation}
    \begin{split}
        \boldsymbol{v} &=
        \begin{bmatrix}
            \symbolroadplane{v}_x + \omega_y h \\
            -\omega_xh                         \\
            w + \symbolroadplane{v}_z
        \end{bmatrix}
        =
        \begin{bmatrix}
            \symbolroadplane{v}_x \\
            0                     \\
            \symbolroadplane{v}_z
        \end{bmatrix} +
        \begin{bmatrix}
            \omega_y h  \\
            -\omega_x h \\
            w
        \end{bmatrix}
        \\
        &= \symbolroadplane{\boldsymbol{v}} + \boldsymbol{\zeta}_{\boldsymbol{v}}
    \end{split}
    \label{eq:vel-trafo}
\end{equation}

Finally, we project the resulting velocity vector into the vehicle frame by applying the rotation matrix $\boldsymbol{R}_z(\beta)$:
\begin{equation}
    \framevehicle{\boldsymbol{v}} = \boldsymbol{R}_z(\beta) \boldsymbol{v} =  \framevehicle{\symbolroadplane{\boldsymbol{v}}} + \boldsymbol{R}_z(\beta) \boldsymbol{\zeta}_{\boldsymbol{v}}
\end{equation}

\subsubsection{Angular Velocity Transformation}

The transformation relies on the expression for the angular velocity established by Rowold et al. \cite{rowold2023OnlineTimeOptimalTrajectory} with the angular velocities with respect to the arc length $\boldsymbol{\Omega}$:
\begin{equation}
    \label{eq:omega_in_vel}
    \begin{split}
        \boldsymbol{\omega} = \begin{bmatrix}
            \omega_x \\
            \omega_y \\
            \omega_z
        \end{bmatrix}
        &=
        \begin{bmatrix}
            \left(\Omega_x \cos{\chi} + \Omega_y \sin{\chi}\right)\dot{s} \\
            \left(\Omega_y \cos{\chi} - \Omega_x \sin{\chi}\right)\dot{s} \\
            \Omega_z \dot{s} + \dot{\chi}
        \end{bmatrix}\hspace{-0.1cm}\text{.}
    \end{split}
\end{equation}

Evaluating this expression under the constraints in \eqref{eq:assumptions} yields the components intrinsic to the planar vehicle model:

\begin{equation}
    \symbolroadplane{\boldsymbol{\omega}} =
    \begin{bmatrix}
        \symbolroadplane{\omega}_x \\
        \symbolroadplane{\omega}_y \\
        \symbolroadplane{\omega}_z
    \end{bmatrix} \stackrel{!}{=} \begin{bmatrix}
        0 \\
        0 \\
        \Omega_z \dot{s} + \dot{\chi}
    \end{bmatrix}
\end{equation}

Inserting this vector back into \eqref{eq:omega_in_vel} yields the transformation for the angular velocity:

\begin{equation}
    \boldsymbol{\omega} =
    \begin{bmatrix}
        0 \\
        0 \\
        \symbolroadplane{\omega}_z
    \end{bmatrix}
    +
    \begin{bmatrix}
        \left(\Omega_x \cos{\chi} + \Omega_y \sin{\chi}\right)\dot{s} \\
        \left(\Omega_y \cos{\chi} - \Omega_x \sin{\chi}\right)\dot{s} \\
        0
    \end{bmatrix} = \symbolroadplane{\boldsymbol{\omega}} + \boldsymbol{\zeta}_{\boldsymbol{\omega}}
\end{equation}

Transforming the angular velocity vector from the velocity frame into the vehicle frame yields:
\begin{equation}
    \framevehicle{\boldsymbol{\omega}}
    = \boldsymbol{R}_z(\beta) \boldsymbol{\omega}
    = \framevehicle{\symbolroadplane{\boldsymbol{\omega}}} + \boldsymbol{R}_z(\beta)\boldsymbol{\zeta}_{\boldsymbol{\omega}}
\end{equation}

\subsubsection{Accelerations}

The acceleration vector of the center of mass is derived from its linear velocity $\boldsymbol{v}$ and angular velocity $\boldsymbol{\omega}$ through the kinematic relation:
\begin{equation}
    \label{eq:acc_com}
    \begin{split}
        \boldsymbol{a} &= \dot{\boldsymbol{v}} + \boldsymbol{\omega} \times \boldsymbol{v}\\
        &=\begin{bmatrix}
            \dot{v}_x + \dot{\omega}_yh + 2\omega_y\dot{h} + \omega_yw + \omega_x\omega_zh
            \\
            -\dot{\omega}_xh - 2\omega_x\dot{h} + \omega_zv_x - \omega_xw + \omega_y\omega_zh
            \\
            \dot{w} + \ddot{h} - \left(\omega^2_x + \omega^2_y\right)h - \omega_yv_x
        \end{bmatrix} \\
    \end{split}
\end{equation}

For operations on a three-dimensional track, spatial gravitational acceleration must be explicitly accounted for. Following the formulation of Rowold et al. \cite{rowold2023OnlineTimeOptimalTrajectory}, we augment the expression for the acceleration of the center of mass with the gravitational components:
\begin{equation}
    \label{eq:acc_com_grav}
    \begin{split}
        \boldsymbol{a} =&
        \begin{bmatrix}
            \dot{v}_x + \dot{\omega}_yh + 2\omega_y\dot{h} + \omega_yw + \omega_x\omega_zh
            \\
            -\dot{\omega}_xh - 2\omega_x\dot{h} + \omega_zv_x - \omega_xw + \omega_y\omega_zh
            \\
            \dot{w} + \ddot{h} - \left(\omega^2_x + \omega^2_y\right)h - \omega_yv_x
        \end{bmatrix}
        + \\ &
        \begin{bmatrix}
            g (\cos{\mu} \sin{\varphi} \sin{\chi} -\sin{\mu} \cos{\chi})
            \\
            g(\sin{\mu} \sin{\chi} + \cos{\mu} \sin{\varphi} \cos{\chi})
            \\
            g (\cos{\mu} \cos{\varphi})
        \end{bmatrix} \\
    \end{split}
\end{equation}

Applying the foundational assumptions of the planar vehicle dynamics model from \eqref{eq:assumptions} yields the reduced expression for a vehicle constrained to a planar surface:

\begin{equation}
    \label{eq:acc-tilde-plane}
    \symbolroadplane{\boldsymbol{a}} =
    \begin{bmatrix}
        \symbolroadplane{a}_x \\
        \symbolroadplane{a}_y \\
        \symbolroadplane{a}_z
    \end{bmatrix} \stackrel{!}{=}
    \begin{bmatrix}
        \dot{v}_x   \\
        \omega_zv_x \\
        \ddot{h} + g
    \end{bmatrix} \\
\end{equation}

Substituting this expression back into the general spatial acceleration in \eqref{eq:acc_com_grav} yields the transformation from the planar acceleration to its three-dimensional counterpart:

\begin{equation}
    \label{eq:acc-trafo}
    \begin{split}
        \boldsymbol{a} =&
        \begin{bmatrix}
            \symbolroadplane{a}_x \\
            \symbolroadplane{a}_y \\
            \symbolroadplane{a}_z
        \end{bmatrix}
        +
        \begin{bmatrix}
            \dot{\omega}_yh + 2\omega_y\dot{h} + \omega_yw + \omega_x\omega_zh
            \\
            -\dot{\omega}_xh - 2\omega_x\dot{h} - \omega_xw  + \omega_y\omega_zh
            \\
            \dot{w} - \left(\omega^2_x + \omega^2_y\right)h - \omega_yv_x
        \end{bmatrix}
        + \\ &
        \begin{bmatrix}
            g (\cos{\mu} \sin{\varphi} \sin{\chi} -\sin{\mu} \cos{\chi})
            \\
            g(\sin{\mu} \sin{\chi} + \cos{\mu} \sin{\varphi} \cos{\chi})
            \\
            g (\cos{\mu} \cos{\varphi} - 1)
        \end{bmatrix} \\
        =& \symbolroadplane{\boldsymbol{a}} + \boldsymbol{\zeta}_{\boldsymbol{a}} + \boldsymbol{\Gamma}
    \end{split}
\end{equation}

Finally, the resulting acceleration vector is rotated by the sideslip angle $\beta$ to project the quantities from the velocity frame into the vehicle frame:

\begin{equation}
    \framevehicle{\boldsymbol{a}}
    = \boldsymbol{R}_z(\beta) \boldsymbol{a}
    = \framevehicle{\symbolroadplane{\boldsymbol{a}}} + \boldsymbol{R}_z(\beta)\left(\boldsymbol{\zeta}_{\boldsymbol{a}}+\boldsymbol{\Gamma}\right)
\end{equation}
\subsubsection{Rotational Accelerations}

As detailed in Appendix~\ref{sec:angular-accelerations}, the transformation relating the angular accelerations of the planar vehicle model to the corresponding spatial variables on the three-dimensional track is given by:
\begin{equation}
    {\framevehicle{\dot{\boldsymbol{\omega}}}} = \framevehicle{\dot{\symbolroadplane{\boldsymbol{\omega}}}} + \boldsymbol{R}_z(\beta)\boldsymbol{\zeta}_{\dot{\boldsymbol{\omega}}}
\end{equation}

\subsection{Vehicle Load Calculation}
\label{sec:load-calc}
Since the simulation must accurately reflect the dynamics of a vehicle operating on a three-dimensional track, computing the forces and moments arising from the combined influence of road geometry and spatial gravitational loads is essential. These spatial loads are omitted in a planar vehicle model and must therefore be evaluated separately and fed back as external perturbation forces and moments acting at the vehicle's center of gravity.

\subsubsection{Forces}

The total force on the vehicle equals its mass multiplied by the acceleration vector:

\begin{equation}
    \framevehicle{\boldsymbol{F}}=m\framevehicle{\boldsymbol{a}}
\end{equation}

The load forces explicitly captured by the unperturbed planar vehicle dynamics model, $\framevehicle{\boldsymbol{\symbolroadplane{F}}}$, are expressed as:
\begin{equation}
    \framevehicle{\symbolroadplane{{\boldsymbol{F}}}} = m\framevehicle{\symbolroadplane{\boldsymbol{a}}}
\end{equation}

The additional forces acting on the vehicle, induced by the three-dimensional road geometry, $\Delta\framevehicle{\boldsymbol{F}}$, are then given by:
\begin{equation}
    \Delta\framevehicle{\boldsymbol{F}} = \framevehicle{\symbolroadplane{\boldsymbol{F}}} - \framevehicle{\boldsymbol{F}} = m \left( \framevehicle{\symbolroadplane{\boldsymbol{a}}} -\framevehicle{\boldsymbol{a}} \right)
\end{equation}

\subsubsection{Moments}

The applied moments are derived using the principle of angular momentum evaluated in a rotating reference frame:

\begin{equation}
    \framevehicle{\boldsymbol{M}} = \frac{d\framevehicle{\boldsymbol{L}}}{dt}\Big|_{\text{inertial}}
    = \framevehicle{\dot{\boldsymbol{L}}}
    + \framevehicle{\boldsymbol{\omega}} \times \framevehicle{\boldsymbol{L}}
\end{equation}

Using $\framevehicle{\boldsymbol{L}} = \boldsymbol{I}\framevehicle{\boldsymbol{\omega}}$, where $\boldsymbol{I} = \mathrm{diag}(I_x, I_y, I_z)$ is the inertia tensor of the modeled vehicle body, we obtain an explicit expression for the moments acting on the vehicle:

\begin{equation}
    \label{eq:euler-formula}
    \begin{split}
        \framevehicle{\boldsymbol{M}} &=
        \begin{bmatrix}
            \framevehicle{M}_x \\
            \framevehicle{M}_y \\
            \framevehicle{M}_z
        \end{bmatrix} =
        \boldsymbol{I}  \framevehicle{\dot{\boldsymbol{\omega}}} + \framevehicle{\boldsymbol{\omega}} \times \left( \boldsymbol{I}\framevehicle{\boldsymbol{\omega}} \right) = \\ &=
        \begin{bmatrix}
            I_x \framevehicle{\dot{\omega}_x} + (I_z - I_y)\framevehicle{\omega_y}\framevehicle{\omega_z}  \\
            I_y\framevehicle{\dot{\omega}_y} + (I_x - I_z)\framevehicle{\omega_x}  \framevehicle{\omega_z} \\
            I_z\framevehicle{\dot{\omega}_z} + (I_y - I_x)\framevehicle{\omega_x}\framevehicle{\omega_y}
        \end{bmatrix}
    \end{split}
\end{equation}

The expanded derivation of the underlying angular accelerations $\dot{\omega}$ is provided in Appendix~\ref{sec:angular-accelerations}.

The moments explicitly captured by the planar vehicle dynamics model, denoted as $\boldsymbol{\symbolroadplane{M}}$, are obtained by applying the assumptions from \eqref{eq:assumptions} to \eqref{eq:euler-formula}:
\begin{equation}
    \framevehicle{\symbolroadplane{\boldsymbol{M}}} = \begin{bmatrix}
        \framevehicle{\symbolroadplane{M}_x} \\
        \framevehicle{\symbolroadplane{M}_y} \\
        \framevehicle{\symbolroadplane{M}_z}
    \end{bmatrix} = \begin{bmatrix}
        0 \\
        0 \\
        I_z  \framevehicle{\dot{\symbolroadplane{\omega}}}_z
    \end{bmatrix}.
\end{equation}

Given $\framevehicle{\omega_z} = \framevehicle{\symbolroadplane{\omega}}_z$, the additional moments induced by the spatial road geometry, $\Delta\framevehicle{\boldsymbol{M}}$, are given by:
\begin{equation}
    \begin{split}
        \Delta\framevehicle{\boldsymbol{M}} & =  \framevehicle{\symbolroadplane{\boldsymbol{M}}} - \framevehicle{\boldsymbol{M}}
        \\
        & = \begin{bmatrix}
            - I_x\framevehicle{\dot{\omega}_x} - (I_z - I_y)\framevehicle{\omega_y}\framevehicle{\omega_z} \\
            - I_y\framevehicle{\dot{\omega}_y} - (I_x - I_z)\framevehicle{\omega_x}\framevehicle{\omega_z} \\
            - (I_y - I_x) \framevehicle{\omega_x}  \framevehicle{\omega_y}
        \end{bmatrix}
    \end{split}
\end{equation}
\section{Validation \& Results}
\label{sec:results}

After presenting our method in the previous sections, this section focuses on its validation.
First, we use synthetically crafted validation tracks to validate our approach in a tailored environment.
Second, we compare our method's results with real-world data. Finally, we analyze the computational efficiency of our method to confirm its suitability for real-time applications.

We collected the real-world data using the full-scale autonomous Dallara AV21 race car used in the Indy Autonomous Challenge \cite{mitchell2024IndyAutonomousChallenge}.
Therefore, all results presented in this section are based on this vehicle, even though our approach is by design compatible with any vehicle model that can be influenced by external forces and moments. Tab.~\ref{tab:vehicle-parameters} lists its relevant parameters, which were approximated from the recorded data.

\begin{table}[tb]
    \centering
    \caption{Vehicle Parameters Used in Simulation}
    \begin{tabular}{c c c}
        \toprule
        \textbf{Parameter} & \textbf{Value} & \textbf{Unit}                \\
        \midrule
        $m$                & \num{800}      & \si{\kilogram}               \\
        $h$                & \num{0.3}      & \si{\meter}                  \\
        $I_x$              & \num{100}      & \si{\kilogram\meter\squared} \\
        $I_y$              & \num{500}      & \si{\kilogram\meter\squared} \\
        $I_z$              & \num{1000}     & \si{\kilogram\meter\squared} \\
        \bottomrule
    \end{tabular}
    \label{tab:vehicle-parameters}
\end{table}

Unless stated otherwise, the simulation results presented in this paper were obtained using a planar point-mass model mirroring the mass and inertial properties of the actual vehicle. It perfectly tracks the road plane reference line in an open-loop simulation. This setup improves the interpretability of the results, which would otherwise be compromised by a vehicle-controller combination introducing additional control errors.
In addition, our comparison with real-world data includes a closed-loop evaluation to demonstrate the validity of our approach for more complex simulation routines.
Within this, we use a motion control algorithm adapted from Wischnewski~et~al.~\cite{wischnewski2022TubeMPCApproachAutonomous} to drive along the Las Vegas Motor Speedway. As the simulation model, we use a planar dual-track model which is openly available~\cite{sagmeister2024AnalyzingImpactSimulation,sagmeister2024OpenCarDynamics}.

\subsection{Synthetic Validation Tracks}

\begin{figure}
    \begin{center}
        \begin{subfigurecustom}
            \begin{minipage}[b]{0.49\columnwidth}
                \centering
                \startsubfigcustom
                \includegraphics[width=\columnwidth]{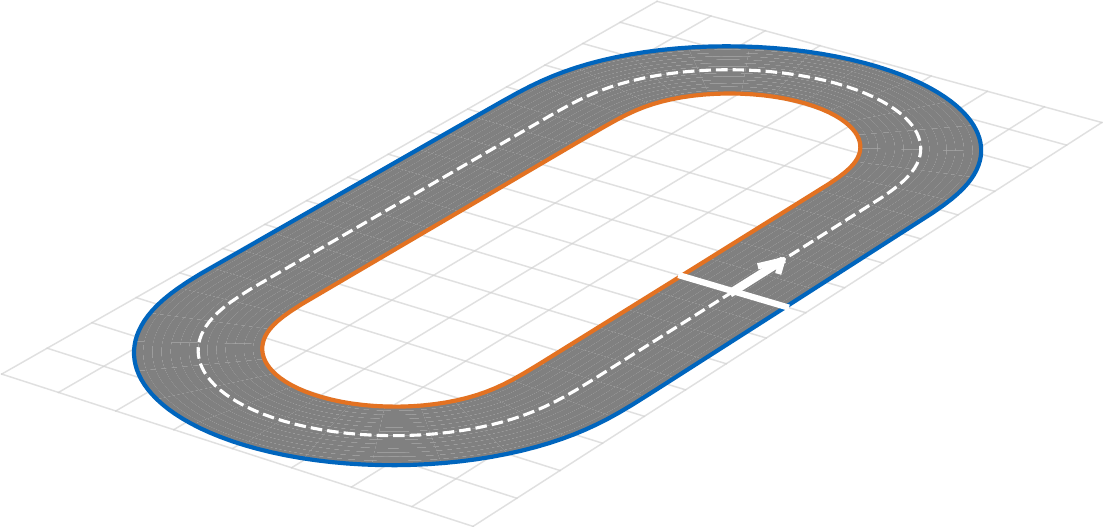}
                \subfigcaptioncustom{\trackflat}
                \label{fig:track-flat}
            \end{minipage}
            \begin{minipage}[b]{0.49\columnwidth}
                \centering
                \startsubfigcustom
                \includegraphics[width=\columnwidth]{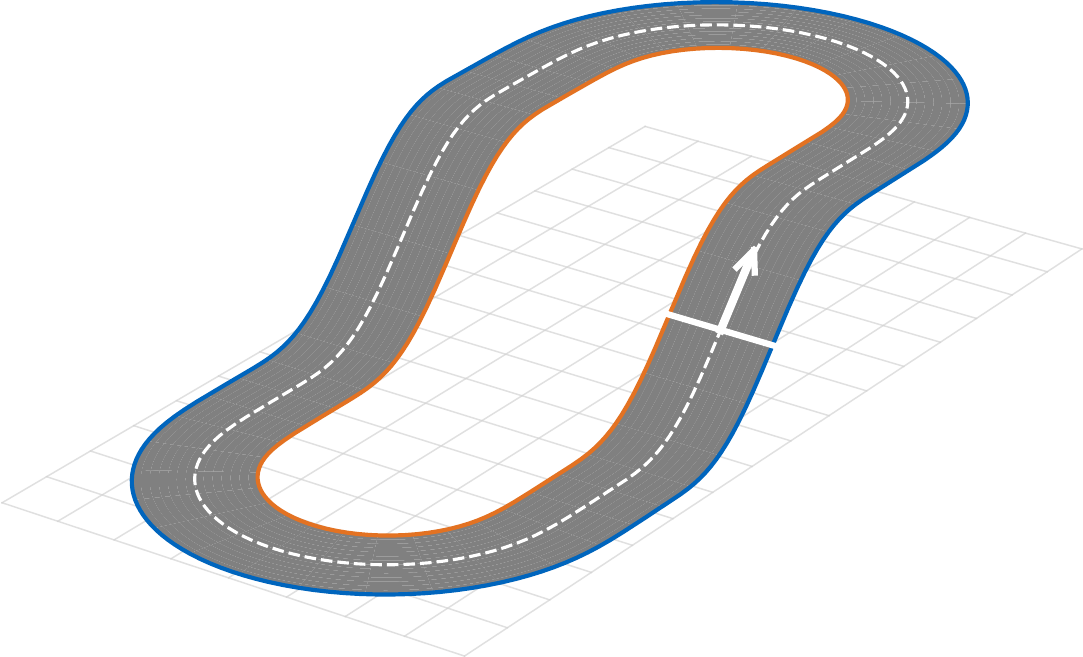}
                \subfigcaptioncustom{\trackelevated}
                \label{fig:track-elevated}
            \end{minipage}
            \par\vspace{0.4cm}

            \begin{minipage}[b]{0.49\columnwidth}
                \centering
                \startsubfigcustom
                \includegraphics[width=\columnwidth]{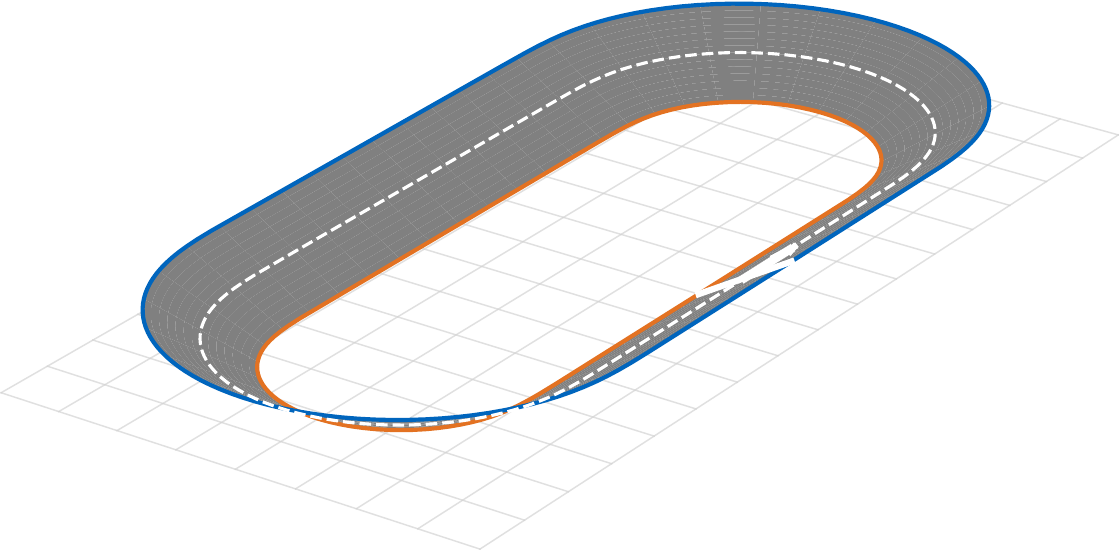}
                \subfigcaptioncustom{\trackbanked}
                \label{fig:track-banked}
            \end{minipage}
            \begin{minipage}[b]{0.49\columnwidth}
                \centering
                \startsubfigcustom
                \includegraphics[width=\columnwidth]{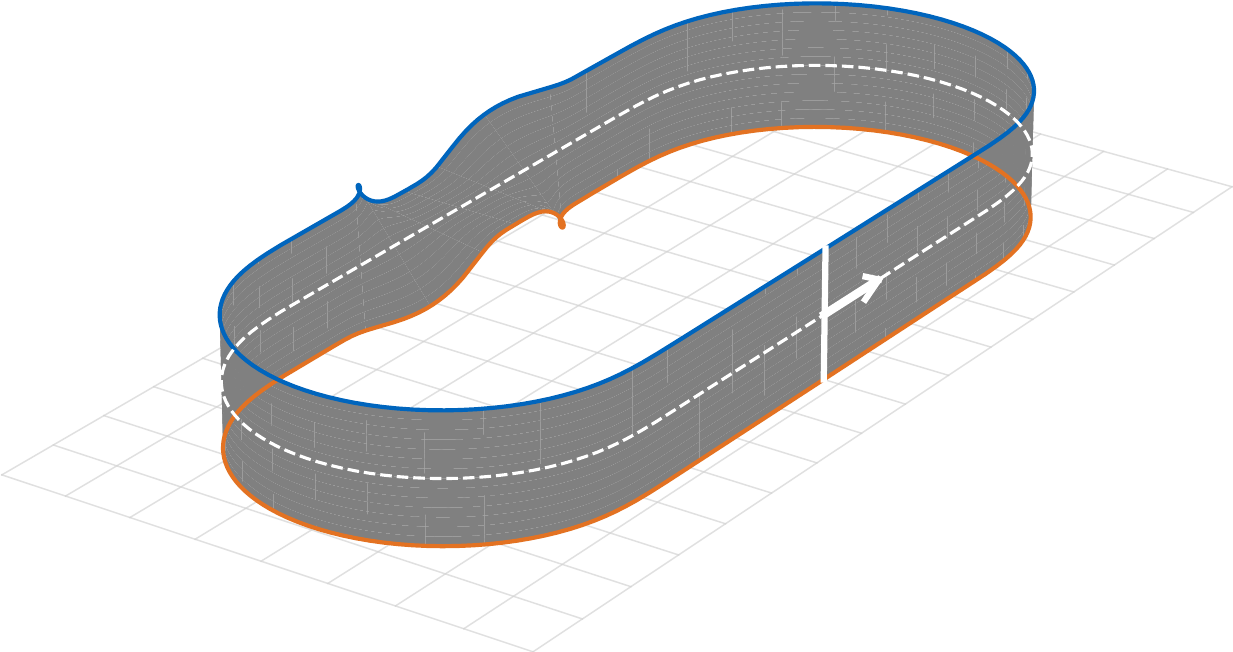}
                \subfigcaptioncustom{\trackverticalbank}
                \label{fig:track-vertical}
            \end{minipage}
        \end{subfigurecustom}
    \end{center}
    \caption{Overview of the synthetic validation tracks. The centerline, which in bird's-eye view is identical among all the tracks, is shown as a dashed white line. The white solid line together with the arrows indicates the starting position and direction of travel.}
    \label{fig:synthetic-validation-tracks} \end{figure}

To validate our method in extremal scenarios, we use four synthetic validation tracks, shown in Fig.~\ref{fig:synthetic-validation-tracks}. The idea behind these tracks is that the bird's-eye view of the centerline is identical across all four tracks. This allows us to fully separate the effects stemming from the track's three-dimensionality from those stemming from its two-dimensional layout. The flat \trackflat{} track serves as the reference, since it is purely two-dimensional.
Each track is designed to isolate, as much as possible, different physical effects caused by the three-dimensionality of the road.
All four tracks share the same starting point and run counterclockwise. The track layout is a simple oval with two straights and two corners. The turn radius smoothly decreases until it reaches a steady minimum of \SI{25}{\meter}. The planar vehicle model moves at a constant speed of \SI{14.1}{\meter\per\second} along the centerline of the track, which results in a lateral acceleration of \SI{8}{\meter\per\second\squared} in the turns on \trackflat{}.
In contrast to \trackflat{}, the \trackelevated{} track adds elevation changes while leaving the corners unchanged.
The \trackbanked{} track has a constant banking angle of \SI{30}{\degree} while keeping the elevation constant.
Finally, \trackverticalbank{} features full vertical banking, except for one straight section where the track twists to a banking angle of \SI{0}{\degree} before returning to the vertical banking.
Although \trackverticalbank{} in particular is not representative of real-world tracks, it proves the validity of our method even in extremal cases and reveals effects that might be hard to identify on more realistic tracks.

\begin{figure}
    \centering
    \inputtikzfig{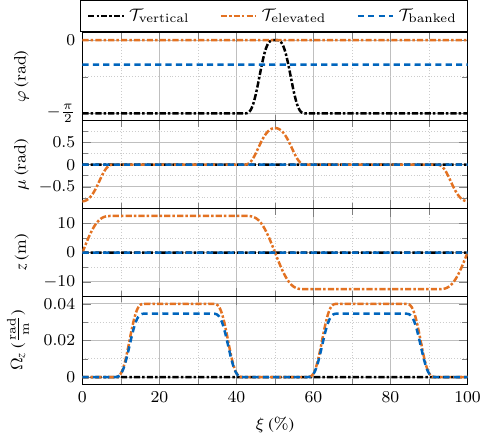}
    \caption{Key characteristics of the synthetic validation tracks. \trackflat{} is omitted as it is purely two-dimensional. Its curvature $\boldsymbol{\Omega}_z$ is identical to the one of \trackelevated{}.}
    \label{fig:characterstics-validation-tracks}
\end{figure}

Fig.~\ref{fig:characterstics-validation-tracks} shows the key characteristics of the validation tracks, such as the curvature within the road plane $\Omega_z$, the banking angle $\varphi$, the slope angle $\mu$, and the elevation $z$.
This already highlights an important factor that is often neglected~\cite{wischnewski2022Tubemodelpredictive,bongard2026RobustNonlinearTrajectory}
when using a planar vehicle dynamics model. Due to the banked turn of \trackbanked{}, the effective curvature within the road plane $\Omega_z$ decreases compared to the unbanked \trackelevated{} track. This means that the planar vehicle effectively needs to follow a smaller curvature on the banked turn than on the unbanked one. For \trackverticalbank{}, the curvature $\Omega_z$ even becomes zero along the entire track.

\begin{figure}
    \centering
    \inputtikzfig{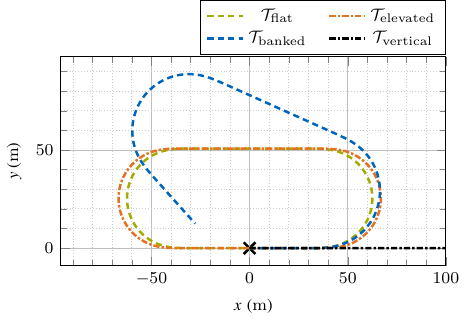}
    \caption{Resulting planar reference line $\frameinertial{\symbolroadplane{\boldsymbol{\mathcal{C}}}}$ for the various validation tracks. Since \trackflat{} is purely two-dimensional, its reference line is identical to the bird's-eye view of the centerline for all the synthetic validation tracks. The initial point for constructing the planar reference line is marked with a black cross.
    }
    \label{fig:ref-lines-validation-tracks}
\end{figure}

Fig.~\ref{fig:ref-lines-validation-tracks} highlights this even further. It shows the reference line in our virtually constructed road plane.
It illustrates that the planar reference lines do not have to be closed, even though the track's centerline is a closed path, as the reference lines for \trackbanked{} and \trackverticalbank{} no longer close. For \trackverticalbank{}, the reference line is even a straight line. This is because the curvature in the road plane is zero due to the fully vertical banking. As a result, a planar vehicle model would simply drive straight when navigating \trackverticalbank{}, highlighting the importance of accounting for changes in road surface geometry when driving on non-flat roads.

Compared to \trackflat{}, the reference line of \trackelevated{} is stretched out on the straight segments. This is because the actual length of the road surface increases with increasing slope angle, given an identical length in the bird's-eye view plane. Since stretching occurs only on the sloped straights, the turn sections are shifted outward but not distorted compared to the flat track.
This highlights the difference between the distance traveled on the road surface and the distance traveled in the bird's-eye view plane.

\begin{figure}
    \centering
    \inputtikzfig{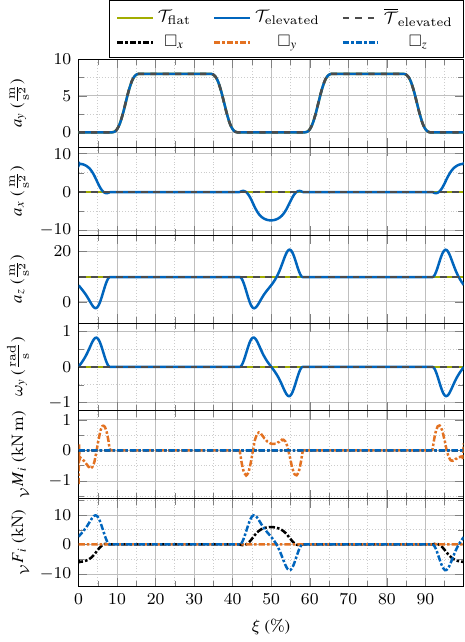}
    \caption{Three-dimensional vehicle states when driving around \trackelevated{} with a constant speed of \SI{14.1}{\meter\per\second}. The signals for the \trackflat{} are given as a reference and correspond to neglecting all three-dimensional effects.
        The states of a planar vehicle dynamics model driving along the reference line of the virtually constructed road plane of \trackelevated{} are depicted as dashed lines. Dash-dotted lines indicate forces and moments acting on the planar vehicle dynamics model due to the three-dimensional road geometry.}
    \label{fig:vehicle-states-validation-track-elevated}
\end{figure}

\begin{figure}
    \centering
    \inputtikzfig{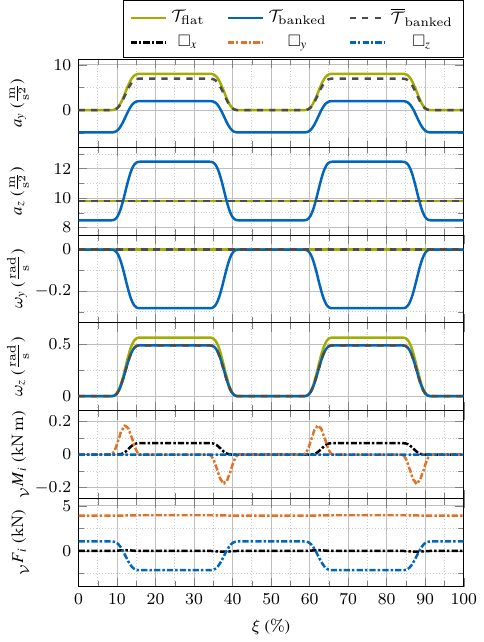}
    \caption{Three-dimensional vehicle states when driving around \trackbanked{} with a constant speed of \SI{14.1}{\meter\per\second}. The signals for the \trackflat{} are given as a reference and correspond to neglecting all three-dimensional effects.
        The states of a planar vehicle dynamics model driving along the reference line of the virtually constructed road plane of \trackbanked{} are depicted as dashed lines. Dash-dotted lines indicate forces and moments acting on the planar vehicle dynamics model due to the three-dimensional road geometry.}
    \label{fig:vehicle-states-validation-track-banked}
\end{figure}

\begin{figure}
    \centering
    \inputtikzfig{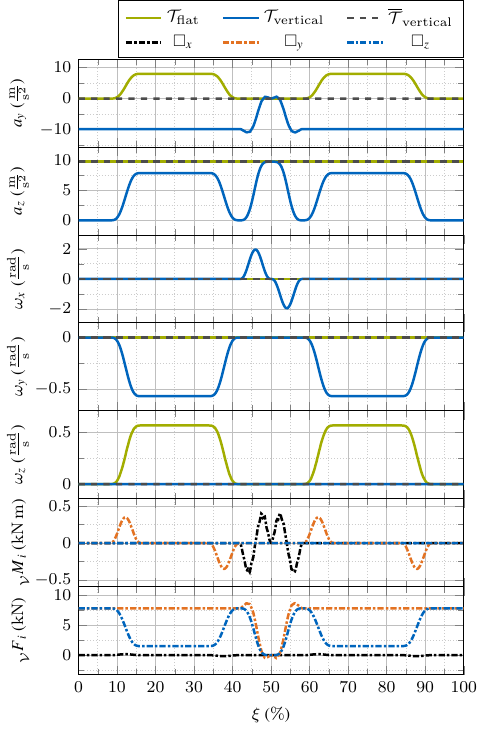}
    \caption{Three-dimensional vehicle states when driving around \trackverticalbank{} with a constant speed of \SI{14.1}{\meter\per\second}. The signals for the \trackflat{} are given as a reference and correspond to neglecting all three-dimensional effects.
        The states of a planar vehicle dynamics model driving along the reference line of the virtually constructed road plane of \trackverticalbank{} are depicted as dashed lines. Dash-dotted lines indicate forces and moments acting on the planar vehicle dynamics model due to the three-dimensional road geometry.}
    \label{fig:vehicle-states-validation-track-vertical}
\end{figure}

While Fig.~\ref{fig:ref-lines-validation-tracks} displays the results of our method projecting the three-dimensional road surface onto a two-dimensional plane, Figs.~\ref{fig:vehicle-states-validation-track-elevated} to~\ref{fig:vehicle-states-validation-track-vertical} present the results of the measurement transformation and the force calculation.
To generate these results, we assume a planar vehicle dynamics model that perfectly follows the reference line in the virtually constructed road plane at a constant speed of \SI{14.1}{\meter\per\second}.
To separately highlight the importance of considering three-dimensional road geometry, the signals for \trackflat{} are included as a reference, illustrating what would happen if these effects were neglected.
Further, these figures show the untransformed states of the planar vehicle dynamics model driving in the virtually constructed road plane, as well as the forces and moments acting on the vehicle due to the road geometry.

For \trackelevated{}, we do not expect any change in lateral acceleration $a_y$ or yaw rate $\omega_z$, since both turns are locally flat. Fig.~\ref{fig:vehicle-states-validation-track-elevated} confirms this, as the signals for $a_y$ and $\omega_z$ are identical to those of \trackflat{}. In contrast, the longitudinal acceleration $a_x$ and the pitch rate $\omega_y$ clearly reflect the ascent and descent of the slope. While $a_x$ primarily reflects the stationary gravitational acceleration due to the slope, $\omega_y$ captures the rotational effect of the gradually changing slope angle. This rotational effect, combined with the vehicle's velocity, leads to compression (increased $a_z$) as the slope angle increases and decompression (decreased $a_z$) as the slope angle decreases. The effect is so pronounced that the vehicle essentially becomes ``weightless'' at the crest of the slope.
Analogously, the forces and moments both show this dynamic compression and decompression phase in $F_z$, while $F_x$ shows the stationary gravitational force when going up and down the hill.

Since the plot uses spatial progress along the bird's-eye-view centerline as its $x$-axis, the stretching of the reference line is not visible. It would, however, be visible when plotting the signals over time. Showing this stretching effect would complicate the comparison between the different tracks, since the signals would be stretched differently for each track, which is why we use the spatial representation. For reference, Fig.~\ref{fig:progress-over-time-validation-tracks} in the appendix shows the progress over time for \trackflat{} and \trackelevated{}, visualizing this effect.

For \trackbanked{}, Fig.~\ref{fig:vehicle-states-validation-track-banked} reveals a clear offset in lateral acceleration $a_y$ between the flat and the banked track. As the lateral force $\framevehicle{F_y}$ indicates, the primary contribution is the stationary component of the gravitational acceleration acting in the $y$-direction due to the banking. However, the figure also reveals the dynamic influence of the road geometry transformation: the planar vehicle dynamics model requires less lateral acceleration $a_y$ and a lower yaw rate $\omega_z$ than when driving on the flat track at the same velocity. For the vertical acceleration $a_z$, we can observe a stationary effect in the form of a reduction in gravitational acceleration in the $z$-direction due to the banking during straight driving. In addition, a dynamic effect is visible in the banked turns, since the vertical acceleration increases, resulting in additional positive vertical loads acting on the vehicle. Further, rotational moments $\framevehicle{M_y}$ act on the vehicle when entering and exiting the banked turns, and a gyroscopic roll moment $\framevehicle{M_x}$ acts throughout the banked turns.

Fig.~\ref{fig:vehicle-states-validation-track-vertical} confirms that our approach yields valid results even in extremal cases. For \trackverticalbank{}, the curvature within the road plane is zero. Therefore, the planar vehicle model drives straight ($\symbolroadplane{a}_y = 0$), resulting in a yaw rate $\omega_z$ of zero. However, there is still lateral acceleration $a_y$ due to the stationary gravitational acceleration that now acts in the $y$-direction. Only on the straight section, where the banking angle drops to zero, does $a_y$ also vanish. Because of the fully vertical banking on most portions of \trackverticalbank{}, $a_z$ does not show the expected gravitational acceleration but instead mirrors the lateral acceleration from the flat track, with the exception of the straight section where the banking goes to zero and $a_z$ shows the gravitational acceleration again. The forces and moments show effects similar to those for \trackbanked{}, but with even more pronounced dynamic effects due to the fully vertical banking. Since the car now only rotates around its $y$-axis, the gyroscopic roll moment $\framevehicle{M_x}$ disappears.

\subsection{Comparison with Real-World Data}

\begin{figure}[!tb]
    \centering
    \includegraphics[width=0.65\columnwidth]{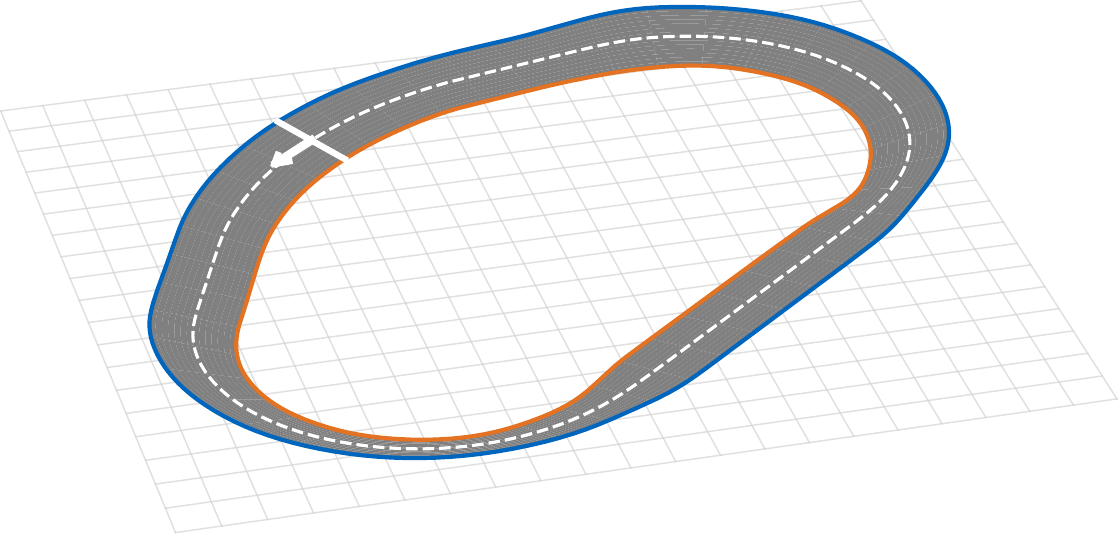}
    \caption{Track layout of the Las Vegas Motor Speedway. The centerline is shown as a dashed white line. The white solid line and the arrow indicate the starting position and direction of travel.}
    \label{fig:vegas-track-layout}
\end{figure}

Having validated our method on synthetic tracks, we now compare its results to real-world data.
We use real-world data collected at a constant velocity of \SI{69.5}{\meter\per\second} on the Las Vegas Motor Speedway. Fig.~\ref{fig:vegas-track-layout} shows its three-dimensional track layout. This racing line also serves as the baseline for all other results presented in the remainder of this section.

\begin{figure}
    \centering
    \inputtikzfig{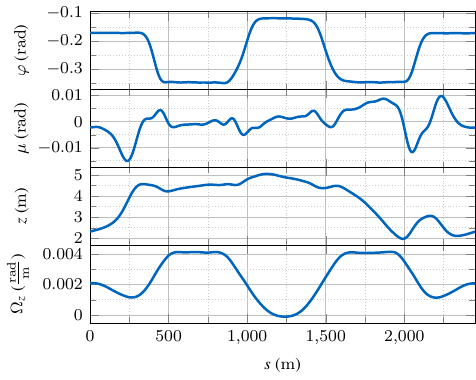}
    \caption{Key characteristics of the Las Vegas Motor Speedway along the racing line used for real-world data collection. The characteristics were computed from the real-world data on low-speed laps using an approach adapted from~\cite{perantoni2015OptimalControlFormula}.}
    \label{fig:vegas-phi-mu-z}
\end{figure}

Fig.~\ref{fig:vegas-phi-mu-z} shows the track's characteristics when following this racing line. It illustrates that while the turns are banked at angles of up to approximately \SI{20}{\degree}, the changes in elevation and slope angles are much smaller.

\begin{figure}
    \centering
    \inputtikzfig{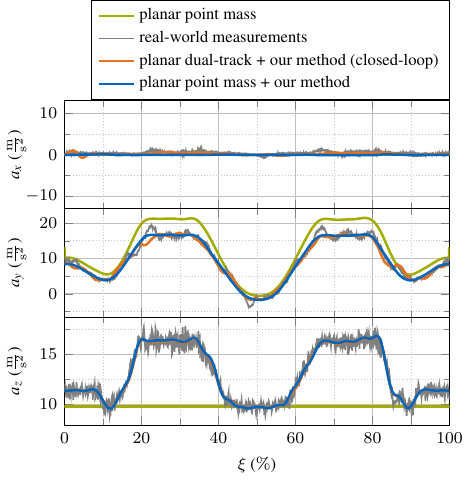}
    \caption{Longitudinal, lateral, and vertical accelerations $a_x$, $a_y$, and $a_z$ on the Las Vegas Motor Speedway at a constant velocity of \SI{69.5}{\meter\per\second}.
        Shown are the real-world measurements, a planar vehicle model following the bird's-eye view reference line of the track (neglecting all three-dimensional effects), our method in the open-loop point-mass setup, and the closed-loop evaluation with the dual-track vehicle model and motion controller introduced above.}
    \label{fig:vehicle-states-real-world-accel}
\end{figure}

\begin{figure}
    \centering
    \inputtikzfig{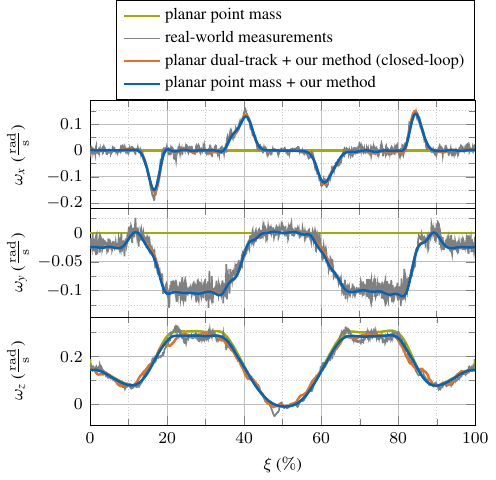}
    \caption{Roll, pitch, and yaw rates $\omega_x$, $\omega_y$, and $\omega_z$ on the Las Vegas Motor Speedway at a constant velocity of \SI{69.5}{\meter\per\second}. Shown are the real-world measurements, a planar vehicle model following the bird's-eye view reference line of the track (neglecting all three-dimensional effects), our method in the open-loop point-mass setup, and the closed-loop evaluation with the dual-track vehicle model and motion controller introduced above.}
    \label{fig:vehicle-states-real-world-rates}
\end{figure}

Figs.~\ref{fig:vehicle-states-real-world-accel} and~\ref{fig:vehicle-states-real-world-rates} show four signals per channel: the recorded real-world data, the planar vehicle model that follows the bird's-eye view reference line of the Las Vegas Motor Speedway and thereby neglects all three-dimensional effects, our method in the open-loop setup described above, and a closed-loop evaluation in which our method drives a dual-track vehicle model via a motion controller.
The real-world data were acquired with an active, closed-loop autonomous driving software stack on the test vehicle, which caused minor velocity fluctuations and lateral oscillations around the targeted reference line, particularly in $a_y$ and $\omega_z$.
The specific oscillation pattern depends on the controller used during data collection, which is no longer reproducible in our current software stack.
However, the closed-loop evaluation is not intended to replicate the exact real-world oscillations; instead, it demonstrates that our method also remains numerically stable and produces physically plausible dynamics in closed-loop simulations.

Fig.~\ref{fig:vehicle-states-real-world-accel} shows the accelerations $a_x$, $a_y$, and $a_z$. While the longitudinal acceleration is barely affected by the small slope angles of the track, the lateral acceleration $a_y$ differs substantially between the planar vehicle dynamics model and the real-world data. This is even more apparent for the vertical acceleration $a_z$, where the planar vehicle dynamics model shows a constant gravitational acceleration of \SI{9.81}{\meter\per\second\squared}, while the real-world data show a considerable increase in vertical acceleration. In turns, the vertical acceleration is between \SI{16}{\meter\per\second\squared} and \SI{17}{\meter\per\second\squared}. Consequently, the resulting vertical forces acting on the vehicle increase the normal load on the tires by over \SI{66}{\percent} relative to the nominal load at standstill, thereby drastically altering the vehicle's dynamic response.

This highlights the importance of considering all influences from the three-dimensional road geometry, rather than accounting only for changes in the orientation of the gravitational force.
Fig.~\ref{fig:vehicle-states-real-world-accel} also shows that by combining the planar vehicle dynamics model with our method, we can accurately reproduce the accelerations measured in the real-world data.

The same applies to the angular velocities. Fig.~\ref{fig:vehicle-states-real-world-rates} shows that while the planar model expectedly does not produce any track-related pitch and roll rates, our method accurately reproduces the collected data.
The yaw rate $\omega_z$ is of particular interest.
Especially in the banked turns, the yaw rate of the planar vehicle dynamics model exceeds the measured value.
Since $\omega_z$ does not have to be transformed, this discrepancy highlights the importance of the geometry transformation using the reference line in the virtually constructed road plane.

Across all signals, the closed-loop evaluation tracks the open-loop result of our method closely, with deviations consistent with the controller's tracking error. As expected, its specific oscillation pattern differs from the real-world recordings, since both the control algorithm and the vehicle response (real-world vehicle vs. simulation model) differ slightly.

To complement the qualitative comparison in Figs.~\ref{fig:vehicle-states-real-world-accel} and~\ref{fig:vehicle-states-real-world-rates} with quantitative metrics, Tab.~\ref{tab:error_metrics} reports the mean error (ME) and mean absolute error (MAE) of each signal against the recorded data, both for the planar baseline and for our method.
For these quantitative results, we filter the recorded data using a fourth-order lowpass Butterworth filter~\cite{butterworth1930TheoryFilterAmplifiers} with a cutoff frequency of \SI{2.0}{\hertz} to mitigate the influence of sensor noise and vibrations.
In addition to MAE, we use the ME because it is invariant to the controller-induced oscillations in the real-world data, which would otherwise heavily influence the MAE. The table shows that our method significantly reduces error relative to the planar baseline, especially for the lateral and vertical accelerations $a_y$ and $a_z$, as well as for the pitch rate $\omega_y$. As already discussed previously, the \SI{82}{\percent} bias reduction on $\omega_z$ deserves special note: since this signal is not modified by the measurement transformation, the improvement originates entirely from the geometry transformation that drives the planar model along the road-plane reference line rather than the bird's-eye view.
\begin{table}[tb]
    \centering
    \caption{Mean error (ME) and mean absolute error (MAE) of the planar baseline and of our method against real-world data recorded on the Las Vegas Motor Speedway at \SI{69.5}{\meter\per\second}. The table shows turns 1 and 2, where the effects of the three-dimensional road geometry are most pronounced. This corresponds to progress from \SI{10}{\percent} to \SI{45}{\percent} along the track.}
    \label{tab:error_metrics}
    \begin{tabular}{llcccc}
        \toprule
                   &                                                 & \multicolumn{2}{c}{ME} & \multicolumn{2}{c}{MAE}                   \\
        \cmidrule(lr){3-4} \cmidrule(lr){5-6}
        Signal     & Unit                                            & Planar                 & Ours                    & Planar & Ours   \\
        \midrule
        $a_x$      & \multirow{3}{*}{\si{\meter\per\second\squared}} & -0.5377                & -0.5264                 & 0.5377 & 0.5264 \\
        $a_y$      &                                                 & 3.2137                 & -0.2381                 & 3.2284 & 0.7046 \\
        $a_z$      &                                                 & -4.2042                & 0.0587                  & 4.2168 & 0.2262 \\
        \midrule
        $\omega_x$ & \multirow{3}{*}{\si{\radian\per\second}}        & -0.0049                & -0.0005                 & 0.0365 & 0.0069 \\
        $\omega_y$ &                                                 & 0.0631                 & -0.0043                 & 0.0635 & 0.0050 \\
        $\omega_z$ &                                                 & 0.0128                 & 0.0023                  & 0.0173 & 0.0106 \\
        \bottomrule
    \end{tabular}
\end{table}

\subsection{Computational Efficiency}

To test the computational efficiency of our method, we benchmarked the C++ implementation on an Intel Core i7-11850H CPU. We drove five laps around the Las Vegas Motor Speedway with a constant speed of \SI{69.5}{\meter\per\second} while executing our algorithm with an update rate of \SI{100}{\hertz}, matching the output rate of our vehicle dynamics model. Our approach showed a mean computation time of \SI{9.5}{\micro\second} per time step. The maximum execution time measured was \SI{64.1}{\micro\second}.

\begin{figure}
    \centering
    \inputtikzfig{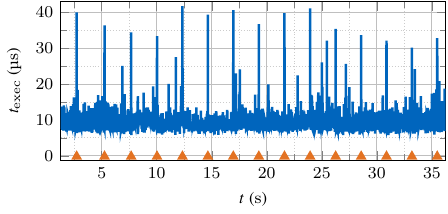}
    \caption{Execution time of our method for 1 lap around the Las Vegas Motor Speedway. The blue line shows the execution time of our method, while the orange triangles indicate when the iterative recalculation of the reference line was performed.}
    \label{fig:runtime}
\end{figure}

Fig.~\ref{fig:runtime} shows the measured execution times for one lap around the Las Vegas Motor Speedway. The blue line shows the execution time of our method, while the orange triangles indicate when the iterative recalculation of the reference line was performed. The iterative recalculation of the reference line takes on average \SI{26.9}{\micro\second}, which is higher than the mean execution time of our method. However, it is still well below the maximum execution time required for real-time applications.

\section{Discussion \& Conclusion}
\label{sec:conclusion}

We presented a modular black-box extension that couples planar vehicle dynamics with three-dimensional road geometry. Experiments on synthetically generated validation tracks demonstrated that the proposed methodology accurately captures physical effects induced by the three-dimensional road geometry, such as dynamic and gravitational loads in the vertical and lateral directions. Furthermore, a comparison with real-world data from an autonomous race car on the Las Vegas Motor Speedway showed that our approach correctly reproduces three-dimensional vehicle states, especially the increase in normal tire load and the altered yaw rates in banked turns compared to a flat road. In the data presented, this reduces the mean absolute error relative to the planar baseline by approximately \SI{95}{\percent} for the vertical and \SI{78}{\percent} for the lateral acceleration, as well as the mean yaw-rate error by \SI{82}{\percent}.

A closed-loop evaluation with a planar dual-track model further confirmed that our method composes with an existing motion control algorithm and remains stable under realistic dynamic deviations from the reference line.

Additionally, our method achieves a mean execution time of \SI{9.5}{\micro\second} per time step, confirming its computational efficiency. This makes our approach suitable for real-time simulation or even sped-up simulations, such as during the training of reinforcement learning agents.
While our method is not intended to be more accurate than fully three-dimensional vehicle dynamics models, it provides a good tradeoff between accuracy and simplicity. It allows us to retain the interpretability of planar vehicle dynamics models while still simulating on three-dimensional roads, which is especially beneficial for developing motion estimation and control algorithms.

Our approach has two principal limitations. First, it uses a rigid-body formulation, treating the forces and moments as acting on the vehicle's center of gravity.
As a result, our method approximates the vehicle's three-dimensional behavior rather than accurately modeling all physical interactions. For example, it does not explicitly account for gyroscopic moments or individual wheel-level suspension kinematics. Nevertheless, our results confirmed that this simplification is valid and that our approach still yields plausible and accurate results.

Second, we use the ribbon approach to model the road geometry. While this enables computational efficiency, it fails to model aspects of the road geometry that cannot be captured in the ribbon representation.
Examples include roads with significant banking changes in the lateral direction, or race tracks with uneven surfaces, e.g., due to bumps or curbs. In future work, this could be addressed by superimposing a more general road representation onto the ribbon. This would allow capturing more complex road geometries while preserving the low computational overhead of our method.

In conclusion, this work demonstrates that the gap between planar simulations and real-world three-dimensional roads can be closed without abandoning planar, and therefore simpler, models, providing a reliable, open-source foundation for autonomous driving development.
\appendices
\section{Derivation of angular accelerations}
\label{sec:angular-accelerations}

The angular velocities are given by:
\begin{equation}
    \boldsymbol{\omega} =
    \begin{bmatrix}
        \omega_x \\
        \omega_y \\
        \omega_z
    \end{bmatrix}
    =
    \begin{bmatrix}
        (\Omega_x \cos{\chi} + \Omega_y \sin{\chi})  \dot{s} \\
        (\Omega_y \cos{\chi} - \Omega_x \sin{\chi})  \dot{s} \\
        \Omega_z \dot{s} + \dot{\chi}
    \end{bmatrix},
\end{equation}

where $\Omega_x$ and $\Omega_y$ are track properties and have to be given from a track file or similar.

From the angular velocities, we can derive the angular accelerations according to the transport theorem:
\begin{equation}
    \left(\frac{d\boldsymbol{\omega}}{dt}\right)_{\mathrm{I}}
    = \dot{\boldsymbol{\omega}}
    + \boldsymbol{\omega} \times \boldsymbol{\omega}
\end{equation}

With $\boldsymbol{\omega} \times \boldsymbol{\omega} = \boldsymbol{0}$ we get:
\begin{equation}
    \boldsymbol{\dot{\omega}} = \begin{bmatrix}
        \dot{\omega}_x \\
        \dot{\omega}_y \\
        \dot{\omega}_z
    \end{bmatrix}
\end{equation}

For $\dot{\omega}_x$ we get:
\begin{align}
    \dot{\omega}_x      & =
    \frac{\mathrm{d}}{\mathrm{d}t}  (\Omega_x  \cos{\chi}  \dot{s})
    + \frac{\mathrm{d}}{\mathrm{d}t}  (\Omega_y  \sin{\chi}  \dot{s}) \label{eq:angular-accel-x}
    \\
    \mathrm{with} \quad & \frac{\mathrm{d}}{\mathrm{d}t}  (\Omega_x  \cos{\chi}  \dot{s}) =
    \Omega_x'  \cos{\chi}  \dot{s}^2
    - \Omega_x  \sin{\chi}  \dot{\chi}  \dot{s}
    + \Omega_x  \cos{\chi}  \ddot{s}
    \\
    \mathrm{and} \quad  & \frac{\mathrm{d}}{\mathrm{d}t}  (\Omega_y  \sin{\chi}  \dot{s}) =
    \Omega_y'  \sin{\chi}  \dot{s}^2
    + \Omega_y  \cos{\chi}  \dot{\chi}   \dot{s}
    + \Omega_y  \sin{\chi}  \ddot{s}
\end{align}

Analogously for $\dot{\omega}_y$ we get:
\begin{equation}
    \begin{split}
        \dot{\omega}_y  = &
        \Omega_y'  \cos{\chi}  \dot{s}^2
        - \Omega_y  \sin{\chi}  \dot{\chi}  \dot{s}
        + \Omega_y  \cos{\chi}  \ddot{s}
        - \\ & \Omega_x'  \sin{\chi}  \dot{s}^2
        - \Omega_x  \cos{\chi}  \dot{\chi}  \dot{s}
        - \Omega_x  \sin{\chi}  \ddot{s}
    \end{split}
    \label{eq:angular-accel-y}
\end{equation}

For $\dot{\omega}_z$ we get:
\begin{equation}
    \dot{\omega}_z = \Omega'_z\dot{s}^2 + \Omega_z\ddot{s} + \ddot{\chi}
\end{equation}

For $\ddot{s}$ we can derive the following expression by differentiating $\dot{s}$:
\begin{align}
    \ddot{s}            & =
    \frac{
        \frac{\mathrm{d}}{\mathrm{d}t}  (v_x  \cos{\chi})  (1 - n  \Omega_z)
        - v_x  \cos{\chi}  \frac{\mathrm{d}}{\mathrm{d}t}  (1 - n  \Omega_z)}
    {
        (1 - n  \Omega_z)^2
    }
    \\
    \mathrm{with} \quad & \frac{\mathrm{d}}{\mathrm{d}t}  (v_x  \cos{\chi}) =
    \dot{v_x}  \cos{\chi}
    - v_x  \sin{\chi}  \dot{\chi}
    \\
    \mathrm{and} \quad  & \frac{\mathrm{d}}{\mathrm{d}t}  ((1 - n  \Omega_z)) =
    - \dot{n}  \Omega_z
    - n  \Omega_z'  \dot{s}
\end{align}

Additionally, we have:
\begin{align}
    \dot{s}    & = \frac{v_x  \cos{\chi}}{1 - n  \Omega_z} \\
    \dot{\chi} & = \omega_z - \Omega_z  \dot{s},           \\
    \dot{v_x}  & = a_x - \omega_y  n  \Omega_x  \dot{s}    \\
    \dot{n}    & = - v_x  \sin{\chi}
\end{align}

$\Omega_x'$ and $\Omega_z'$ are track properties that have to be given from a track file or similar.

With the assumptions from \eqref{eq:assumptions} for the planar vehicle dynamics model the equations evaluate to:
\begin{equation}
    \dot{\symbolroadplane{\boldsymbol{\omega}}} =
    \begin{bmatrix}
        \dot{\symbolroadplane{\omega}}_x \\
        \dot{\symbolroadplane{\omega}}_y \\
        \dot{\symbolroadplane{\omega}}_z
    \end{bmatrix} \stackrel{!}{=}
    \begin{bmatrix}
        0 \\
        0 \\
        \Omega'_z\dot{s}^2 + \Omega_z\ddot{s} + \ddot{\chi}
    \end{bmatrix}
\end{equation}

Inserting these expressions for the angular accelerations into \eqref{eq:acc_com_grav} yields the following transformation for the angular acceleration of the center of mass:

\begin{equation}
    \dot{\boldsymbol{\omega}} =
    \begin{bmatrix}
        0 \\
        0 \\
        \dot{\symbolroadplane{\omega}}_z
    \end{bmatrix} + \begin{bmatrix}
        \omega_x \\
        \omega_y \\
        0
    \end{bmatrix} = \dot{\symbolroadplane{\boldsymbol{\omega}}} + \boldsymbol{\zeta}_{\dot{\boldsymbol{\omega}}}
\end{equation}
$\omega_x$ and $\omega_y$ correspond to the expressions from \eqref{eq:angular-accel-x} and \eqref{eq:angular-accel-y}.

These angular accelerations are rotated by the sideslip angle $\beta$ to get the angular accelerations in the vehicle frame:
\begin{equation}
    {\framevehicle{\dot{\boldsymbol{\omega}}}} = \boldsymbol{R}_z(\beta)\dot{\boldsymbol{\omega}} = \framevehicle{\dot{\symbolroadplane{\boldsymbol{\omega}}}} + \boldsymbol{R}_z(\beta)\boldsymbol{\zeta}_{\dot{\boldsymbol{\omega}}}
\end{equation}

\section{Progress Over Time on Validation Tracks}
\label{sec:progress-over-time-validation-tracks}

\begin{figure}[!h]
    \centering
    \inputtikzfig{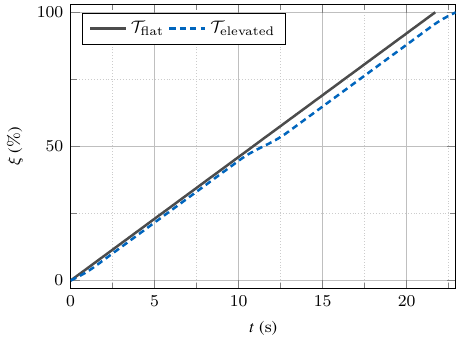}
    \caption{Progress over time for \trackflat{} and \trackelevated{}. The plot visualizes the temporal stretching between tracks that is hidden when signals are plotted over spatial progress.}
    \label{fig:progress-over-time-validation-tracks}
\end{figure}

\FloatBarrier
\section*{Acknowledgment}
Author contributions:
Simon Sagmeister, as the first author, designed the structure of the article and contributed
essentially to the conceptualization, the development, and the implementation of the presented framework.
Phillip Pitschi contributed to the paper writing and the framework's development and implementation.
Nico Haja implemented a first prototype of the framework and supported the paper writing.
Markus Lienkamp made an essential contribution to the concept of the research project. He revised the paper critically for important intellectual content. Markus Lienkamp gives final approval for the version to be published and agrees to all aspects of the work. As a guarantor, he accepts responsibility for the overall integrity of the paper.
AI tools (Claude Opus 4.7, Gemini 3.1 Pro) were used exclusively for editorial, code, and documentation refinement of author-provided input.
\pagebreak

\end{document}